\pdfoutput=1

\documentclass[11pt]{article}

\usepackage[final]{acl}

\usepackage{amsfonts}       
\usepackage{amsmath}

\usepackage{caption}
\usepackage{booktabs} 
\usepackage{times}
\usepackage{latexsym}
\usepackage{subcaption}
\usepackage{multirow}
\usepackage{colortbl} 
\usepackage{xcolor}
\usepackage[most]{tcolorbox}
\usepackage[T1]{fontenc}

\usepackage[utf8]{inputenc}

\usepackage{microtype}

\usepackage{inconsolata}

\usepackage{graphicx}
\newtheorem{definition}{Definition}

\usepackage[capitalize]{cleveref}
\crefname{table}{Table}{Tabs.}
\crefname{figure}{Fig.}{Figs.}
\Crefname{section}{Section}{Sections}
\Crefname{table}{Table}{Tables}
\Crefname{assumption}{Assumption}{Assumptions}
\crefname{algorithm}{Algorithm}{Algorithms}

\title{CausalVLBench: Benchmarking Visual Causal \\ Reasoning in Large Vision-Language Models}

\author{Aneesh Komanduri, Karuna Bhaila, Xintao Wu\\
  Department of Electrical Engineering and Computer Science \\
  University of Arkansas \\
  \texttt{\{akomandu, kbhaila, xintaowu\}@uark.edu}}

\begin{document}
\maketitle
\begin{abstract}
Large language models (LLMs) have shown remarkable ability in various language tasks, especially with their emergent in-context learning capability. Extending LLMs to incorporate visual inputs, large vision-language models (LVLMs) have shown impressive performance in tasks such as recognition and visual question answering (VQA). Despite increasing interest in the utility of LLMs in causal reasoning tasks such as causal discovery and counterfactual reasoning, there has been relatively little work showcasing the abilities of LVLMs on visual causal reasoning tasks. We take this opportunity to formally introduce a comprehensive causal reasoning benchmark for multi-modal in-context learning from LVLMs. Our \textit{CausalVLBench} encompasses three representative tasks: causal structure inference, intervention target prediction, and counterfactual prediction. We evaluate the ability of state-of-the-art open-source LVLMs on our causal reasoning tasks across three causal representation learning datasets and demonstrate their fundamental strengths and weaknesses. We hope that our benchmark elucidates the drawbacks of existing vision-language models and motivates new directions and paradigms in improving the visual causal reasoning abilities of LVLMs. We make our code and data available at \url{https://github.com/Akomand/CausalVLBench}.

\end{abstract}

\section{Introduction}

With a growing emphasis on developing pre-trained models that emulate human reasoning and thinking patterns, a wide range of literature has also focused on assessing these models on complex reasoning tasks~\cite{nie2023moca, mitchell2023comparing}. As human intelligence is often hallmarked by causal reasoning, i.e., the ability to distinguish cause and effect, researchers have also prioritized evaluation of pre-trained models on causal inference tasks~\cite{zhang2023understanding,kiciman2023causal,jin2023cladder}. The evaluation tasks vary among causal effect inference~\cite{jin2023cladder}, causal discovery~\cite{jiralerspong2024efficient, vashishtha2023causal}, and counterfactual reasoning. However, these works mostly focus on analyzing the performance of LLMs, and there has been comparatively little work in evaluating the causal reasoning capability of large visio-linguistic systems.       

Leveraging breakthroughs in contrastive language-image pretraining approaches such as CLIP~\cite{radford2021learning}, LVLMs have sparked new research questions about the capabilities of language models augmented with visual information. Often constructed by integrating a CLIP-like vision encoder in the LLM architecture and aligning the two modalities via projection, LVLMs have shown tremendous potential in tasks such as recognition, grounding, and VQA~\cite{li2025survey}. Recently,~\citet{zong2025vlicl} benchmarked LVLMs on diverse tasks with multi-modal in-context learning (ICL), and ongoing efforts are being made to incorporate and improve reasoning in models such as DeepSeek R1~\cite{guo2025deepseek} and OpenAI o3. Similar to causal evaluations on LLMs, it is important to examine the causal reasoning ability of LVLMs, as building language and vision agents capable of reasoning and planning is paramount to the reliable usage of LLMs and LVLMs in real-world scenarios~\cite{gkountouras2025language}. However, it is significantly more challenging for AI systems to learn causal relationships from high-dimensional data such as images that consist of complex causal dependencies in physical and dynamical systems ~\cite{scholkopf_toward_2021, komanduri2024from}.

In this work, we investigate the causal reasoning capabilities of LVLMs. Previously,~\citet{chen-etal-2024-cello} evaluated the performance of LVLMs on causally-motivated VQA tasks where the causal relationships were defined in terms of observable interactions extracted from the scene graph representation of images. However, their evaluation focused on scene-specific relations originating from human-object interactions. In contrast, we focus on the ability of LVLMs to perform formal visual causal reasoning with systems described by deterministic causal mechanisms (e.g., light position causes a change in shadow length in~\cref{pendulum_structure}). We formulate the visual causal reasoning task as the ability of LVLMs to disentangle causal variables and reason about their relationships, a fundamental goal of causal representation learning~\cite{scholkopf_toward_2021}. We further evaluate \textit{visual} causal reasoning capabilities in LVLMs under zero-shot and few-shot learning settings. Assessing the causal reasoning capability of LVLMs in this setting can have significant implications for building robust AI systems in diverse domains.

Our contributions are: (1) We formulate causal reasoning in LVLMs as inferring causal mechanisms from visual cues. (2) We construct a benchmark, \textit{CausalVLBench}, encompassing three representative tasks: \textbf{causal structure inference}, \textbf{intervention target prediction}, and \textbf{counterfactual prediction} to evaluate the visual causal reasoning capabilities of LVLMs under zero and few shot settings. (3) We study the effect of prompting without the causal graph, demonstration selection, and zero-shot chain-of-thought on the performance of LVLMs. (4) We conduct rigorous empirical evaluations on several LVLMs to assess their performance in understanding the underlying causal mechanisms in physical systems. We primarily experiment with open-source models to maintain transparency and reproducibility. Our experiments indicate that formal causal reasoning while incorporating text and image modalities is challenging for current state-of-the-art LVLMs. We hope our evaluation elucidates the shortcomings of existing models and motivates the development of new training paradigms to promote causal reasoning in large multi-modal models~\cite{vashishtha2024teaching, rajendran2024learning}.

\section{Related Work}
\label{app:related_work}
\paragraph{Causality and LLMs.} Recent studies suggest that LLMs can answer $\mathcal{L}_1$ (observational) questions to a great degree, but struggle to answer $\mathcal{L}_2$ (interventional) and $\mathcal{L}_3$ (counterfactual) questions~\cite{zhang2023understanding}.~\citet{zecevic2023causal} studied the causal reasoning capabilities of large language models and conjectured that they are ``causal parrots'' that may only be learning causal facts from the training data and are not causally reasoning. The authors proposed the notion of \textit{correlation of causal facts} as exploiting a loophole in Pearl's Causal Hierarchy Theorem (CHT) to seemingly talk causality. That is, LLMs may simply be learning correlations about causal facts embedded in the training distribution.

LLMs have been evaluated on a range of different causal inference tasks. ~\citet{jin2023cladder} developed the CLADDER dataset to test the ability of LLMs for the causal effect estimation task and concluded that LLMs perform quite poorly on such tasks. Several works explore the usability of LLMs in causal structure learning tasks by utilizing them as domain experts~\cite{vashishtha2023causal, feder2023causalstructure, romanou2023crab, cohrs2023large, jiralerspong2024efficient}. Another direction probed the interpretability aspect of LLMs from the latent space~\cite{rohekar2023causal, park2023the, rajendran2024learning}. 

\paragraph{Benchmark Evaluation.} Causal evaluation benchmarks have spanned various domains and modalities. Recently, there have been several benchmarks focusing only on the text modality with tasks including mathematical reasoning ~\citep{wang2024causalbench}, common-sense causality ~\citep{miliani-etal-2025-explica}, and causal effect estimation ~\citep{jin2023cladder}. Extending to the vision-language setting, ~\citet{chen-etal-2024-cello} evaluate vision-language models on causal inference tasks. However, their approach focuses on drawing causality from scene graphs representing human-object interactions. Similarly, ~\citet{li-etal-2025-multimodal-causal} propose a multimodal causal reasoning benchmark but still focus on high-level causality. ~\citet{liu2025causal3dcomprehensivebenchmarkcausal} proposed a benchmark 3D dataset for causal reasoning tasks and evaluate a few tasks using closed-source large vision-language models. In contrast, we focus on causal relationships that have a physical and deterministic interpretation (i.e., physical mechanisms) and assess the formal causal reasoning capabilities of pretrained large vision-language models on novel tasks beyond causal structure inference to include intervention target prediction and counterfactual reasoning.

\section{Preliminaries}

\paragraph{Causal Inference.} Our work primarily relies on the causality framework by~\citet{Pearl09} that mathematically formalizes the reasoning of cause and effect. The framework consists of a three-level hierarchy of causal inference referred to as the \textit{Ladder of Causation}~\cite{PearlMackenzie18} or Pearl's Causal Hierarchy (PCH)~\citep{PCH}. 

\begin{itemize}
    \item \textbf{Rung 1}, or $\mathcal{L}_1$, refers to statistical associations among random variables and involves reasoning about joint and conditional distributions. This rung describes questions such as \textit{``How often should I take medication when I am sick?''} 
    \item \textbf{Rung 2}, or $\mathcal{L}_2$, enables interventions on variables to reason about their effects. We can perform an intervention on a random variable $X$ via the do operator (i.e., \textbf{do}$(X=x)$). This rung describes questions such as \textit{``If I take the medication, will my sickness be cured?''}
    \item \textbf{Rung 3}, or $\mathcal{L}_3$, deals with counterfactual reasoning (i.e., ``what if?'' questions) to imagine alternative scenarios in which the world could have been different. This rung describes questions such as \textit{``Would my sickness have been cured if I had taken the medication?''}
\end{itemize}
Reasoning about all of these quantities together requires the Structural Causal Model (SCM) formalism from~\cite{Pearl09} as defined below.
\begin{definition} \normalfont
A structural causal model (SCM) is formally defined by a tuple $\langle Z, U, F\rangle$, where $Z$ is the set of $n$ endogenous variables, $U$ is the set of $n$ exogenous noise variables, and $F$ is a set of structural equations of the form $Z_i = f_i(Z_{pa_i}, U_i)$, where $Z_{pa_i}$ are $Z_i$'s causal parents. The conditionals $P(Z_i | Z_{pa_i})$ define the conditional distribution of $Z_i$ given its parents. The joint observational distribution can then be factorized as follows:
\begin{equation}
    P(Z_1, \dots, Z_n) = \prod_{i=1}^n P(Z_i | Z_{pa_i})
\end{equation}
The observational $(\mathcal{L}_1)$, interventional $(\mathcal{L}_2)$, and counterfactual $(\mathcal{L}_3)$ distributions entailed by the SCM form a hierarchy in the sense that $\mathcal{L}_1 \subset \mathcal{L}_2 \subset \mathcal{L}_3$, where each level encodes richer information that the previous level cannot express.
\end{definition}

\paragraph{In-Context Learning.} Given a pretrained large vision-language model $M_{\theta}$, a text instruction $I$, and some support set $S = (X_i, Y_i)$ of query examples $X$ along with the corresponding answer $Y$ (i.e., demonstrations), and a test query $X^*$, the goal of in-context learning is to estimate the following 
\begin{equation}
    p_{\theta}(Y^* | I, X^*, S)
\end{equation}
In the context of vision-language models, $X$ takes the form of images with a text prompt, and the output $Y$ is generated text.

\begin{figure*}
    \centering
    \includegraphics[width=\linewidth]{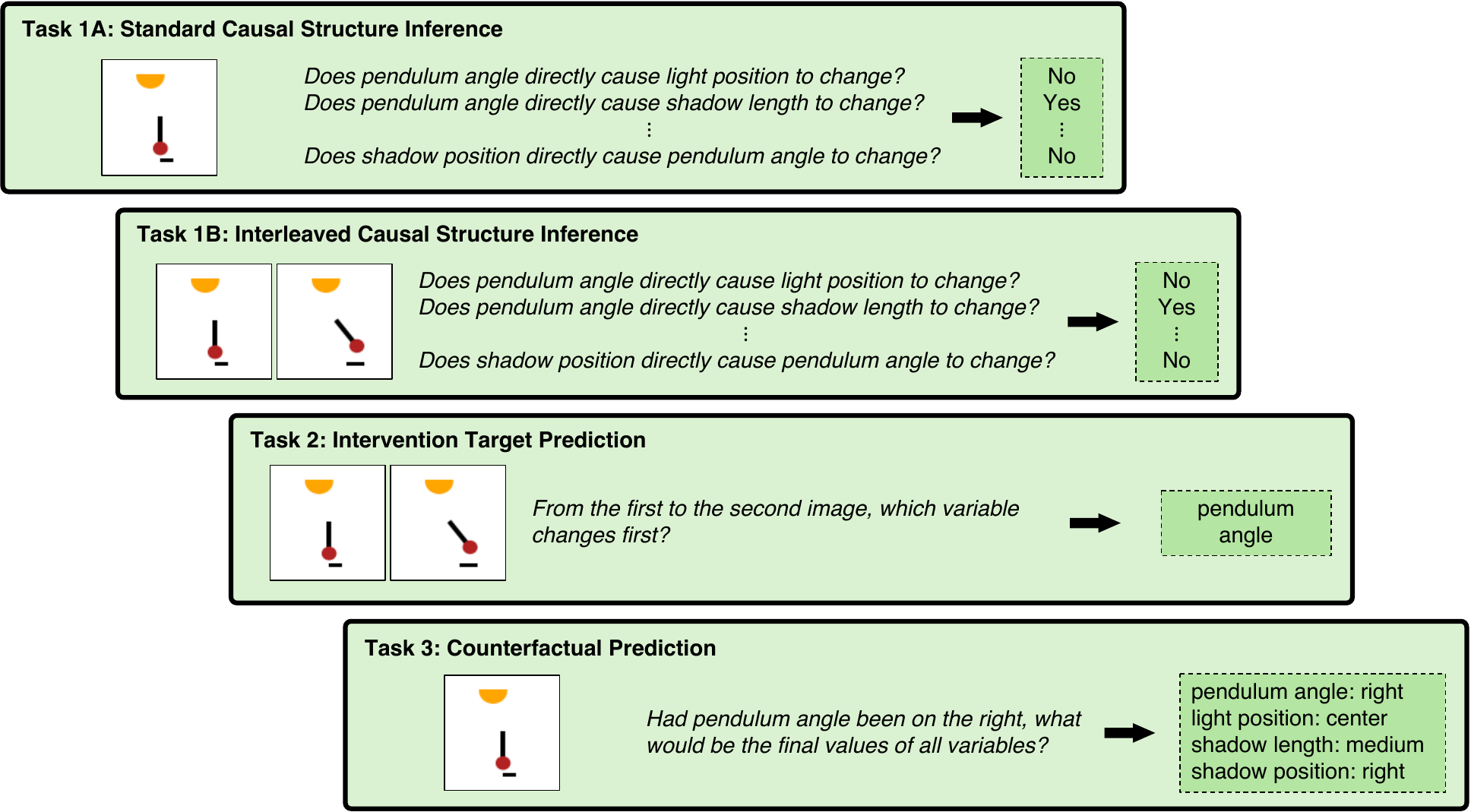}
    \caption{Different causal reasoning tasks including causal structure inference (standard and interleaved), intervention target prediction, and counterfactual prediction.}
    \label{fig:tasks_illustration}
\end{figure*}

\section{CausalVLBench}
In this paper, we study the causal reasoning problem in the extended setting of large vision-language models. We propose three main causal reasoning tasks: causal structure inference, intervention target prediction, and counterfactual prediction.

Let $X = (V, Q)$ be decomposed into a vision and text query. Let $Z = \{Z_1,\dots, Z_n\}$ be the set of causal variables that govern the system shown in the image and also described in the system prompt $I$. For all task formulations, we use $\textbf{\texttt{prompt}}(\cdot)$ to represent a function that converts an input containing a set of causal variables and/or a causal graph to a suitable text prompt.

\begin{figure*}[ht]
    \centering
    \begin{subfigure}[t]{0.3\textwidth}
        \centering
        \includegraphics[height=2.3in]{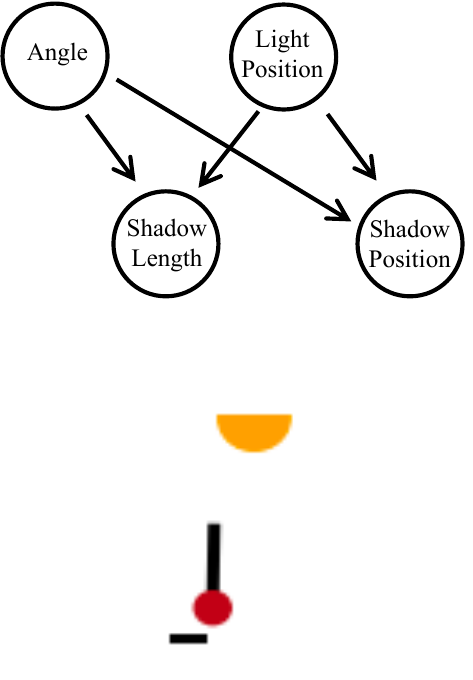}
        \caption{Pendulum}
        \label{pendulum_structure}
    \end{subfigure}~ 
    \begin{subfigure}[t]{0.3\textwidth}
        \centering
        \includegraphics[height=2.3in]{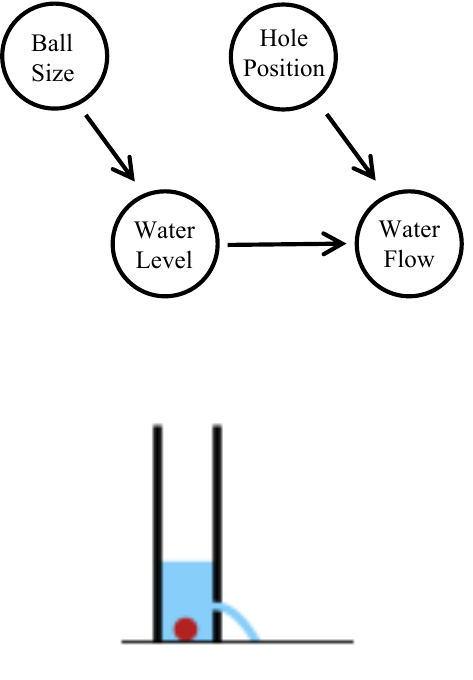}
        \caption{Water Flow}
        \label{flow_structure}
    \end{subfigure}~
    \begin{subfigure}[t]{0.3\textwidth}
        \centering
        \includegraphics[height=2.3in]{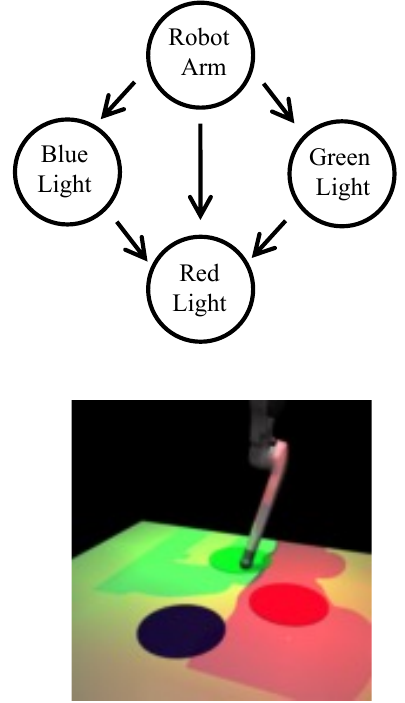}
        \caption{Causal Circuit}
        \label{circuit_structure}
    \end{subfigure}
    \caption{Causal system of all datasets used in evaluations}
    \label{fig:examples_datasets}
\end{figure*}

\paragraph{Task 1: Causal Structure Inference.} Causal discovery refers to learning causal structure from observational or interventional data~\cite{vowels2021d}. In this task, we prompt the LVLM to infer the causal graph from input image(s) and text. Corresponding to causal discovery with observational and interventional data, we formulate two unique settings to evaluate the ability of vision-language models in deriving causal relationships from respective contexts.  
\begin{itemize}
    \item \textbf{Task 1A: Standard Causal Structure Inference.} Given a \textit{single image} and an instructional prompt providing a high-level description of the variables of interest in the image, we prompt the LVLM to infer the causal structure among the given variables through a series of Yes/No questions.
    \item \textbf{Task 1B: Interleaved Causal Structure Inference.} Given an \textit{image pair} and an instructional prompt describing the variables, we prompt the LVLM to infer changes between the images and provide the causal structure among the variables through a series of Yes/No questions. We prompt the LVLM with image pairs to simulate the interventional data scenario where the second image shows the result of an intervention performed on the initial system depicted in the first image.
\end{itemize}

Formally, given input image $V$ (for single image) or a pair of images $V = \{V_{\text{before}}, V_{\text{after}}\}$ (for interleaved), a set of causal variables $Z$, and LVLM $M_{\theta}$, the goal is to infer the causal graph $G = (Z, E)$ where $E$ is the set of directed edges such that $(Z_i \to Z_j) \in E$ indicates that $Z_i$ is a direct cause of $Z_j$. Now, for each pair $(Z_i, Z_j)$, where $i \neq j$, we construct a query $Q_{ij} = \textbf{\texttt{prompt}}(Z_i, Z_j)$ corresponding to the question ``Does $Z_i$ directly cause $Z_j$ to change?'' Then, we have the following binary output from the LVLM:
\begin{equation}
    \hat{Y}_{ij} = M_{\theta}(I, V, Q_{ij})
\end{equation}
where $\hat{Y}_{ij}\in$ \{Yes, No\} and $I$ is the system prompt for the causal structure inference task containing the description of causal variables $Z$. Then, we can construct an adjacency matrix $A$ with entries $\hat{A}_{ij} = \mathbb{I}[\hat{Y}_{ij}] \in \{0, 1\}$. The edge set of the inferred graph $\hat{G} = (Z, \hat{E})$ can be obtained as 
\begin{equation}
    \hat{E} = \{(Z_i, Z_j) \in Z \times Z \mid \hat{A}_{ij} = 1\}
\end{equation}
Similar to traditional causal discovery algorithms that rely on observational or interventional data, our paradigm relies on simple observation or interventional pairs as an inductive bias to identify the causal structure. Figure 1(a) and (b) show an example of Task 1A and Task 1B, respectively.

\paragraph{Task 2: Intervention Target Prediction.} Recent work has shown that a learning paradigm with access to interventional data can be a sufficient signal to recover causal relationships from data~\cite{ahuja_interventional, brehmer2022weakly, zhang2023identifiability}. Inspired by this paradigm, we formulate the task of inferring the original variable that was intervened upon given interventional input data.

Formally, given an image pair, depicting before and after an intervention has affected a system, we propose the task of predicting the source intervention that caused all changes, often referred to as the \textit{intervention target}~\cite{CITRIS}. Suppose we are given a fixed causal graph $G = (Z, E)$, a pair of images $(V_{\text{before}}, V_{\text{after}})$ where a variable from $\{Z_1, \dots, Z_n\}$ was intervened upon, and a query $Q$ = ``From the first to the second image, which variable changes first?''. Then, we have the following output from the LVLM:
\begin{equation}
    \hat{Y} = M_{\theta}(I, \{V_{\text{before}}, V_{\text{after}}\}, Q)
\end{equation}
where $\hat{Y}$ is the predicted intervened target and $I$ is a system prompt for the intervention target prediction task containing the description of causal variables $Z$ and their relationships $G$. This task requires careful reasoning to ensure that the intentional change was not a downstream causal effect.

\paragraph{Task 3: Counterfactual Prediction.} To evaluate the capability of LVLMs to infer causal mechanisms, an important task is counterfactual reasoning. Given an image and a description of the current state of all high-level variables of interest that appear in the image, we prompt an LVLM to infer what the state of all variables would be had a specific intervention taken place. 

Given an image $V$, a fixed causal graph $G = (Z, E)$, and initial variable assignments $\{z_1, \dots, z_n\}$, the goal is to predict the values of all variables $Z_1, \dots, Z_n$ had an intervention \textbf{do}$(Z_i = z_i^*)$ been performed. The query can be represented as $Q = \textbf{\texttt{prompt}}(\{z_1, \dots, z_n\}, \textbf{do}(Z_i = z_i^*))$. We have the following output from the LVLM:
\begin{equation}
    \hat{z} = M_{\theta}(I, V, Q)
\end{equation}
where $I$ is the system prompt for the counterfactual prediction task containing the description of causal variables $Z$ and their relationships $G$ and $\hat{z} = \{\hat{z}_1, \dots, \hat{z}_n\}$ are the LVLM predicted final counterfactual states.

\begin{table*}[htb]
    \renewcommand\arraystretch{1.0}
    \centering
    \caption{Results for \textbf{Task 1A: Standard Causal Structure Inference} and \textbf{Task 1B: Interleaved Causal Structure Inference} task under Zero-Shot setting.}
    \label{tab:causal_structure_results}
    \setlength{\tabcolsep}{2pt}
    \resizebox{\textwidth}{!}{ 
    \begin{tabular}{lcccccccccccc}
    \toprule
         \multirow{4}{*}{\bf{Model}} & \multicolumn{4}{c}{\textbf{Pendulum}} & \multicolumn{4}{c}{\textbf{Water Flow}} & \multicolumn{4}{c}{\textbf{Causal Circuit}} \\ 
         \cmidrule(lr){2-5} \cmidrule(lr){6-9} \cmidrule(lr){10-13}
         & \multicolumn{2}{c}{\textbf{Standard}} & \multicolumn{2}{c}{\textbf{Interleaved}} 
         & \multicolumn{2}{c}{\textbf{Standard}} & \multicolumn{2}{c}{\textbf{Interleaved}} 
         & \multicolumn{2}{c}{\textbf{Standard}} & \multicolumn{2}{c}{\textbf{Interleaved}} \\
         \cmidrule(lr){2-3}\cmidrule(lr){4-5}\cmidrule(lr){6-7}\cmidrule(lr){8-9}\cmidrule(lr){10-11}\cmidrule(lr){12-13}
         & \textbf{SHD} & \textbf{Acc} & \textbf{SHD} & \textbf{Acc} 
         & \textbf{SHD} & \textbf{Acc} & \textbf{SHD} & \textbf{Acc} 
         & \textbf{SHD} & \textbf{Acc} & \textbf{SHD} & \textbf{Acc} \\ 
         \midrule
         LLaVA-OneVision-7B   & $1.2_{0.01}$ & $89.9_{0.06}$ &  $1.7_{0.02}$ & $85.2_{0.14}$ & $2.8_{0.01}$ & $76.3_{0.09}$  & $3.0_{0.00}$ & $75.0_{0.00}$   & $4.4_{0.03}$ & $62.4_{0.24}$ & $3.2_{0.01}$ & $73.4_{0.10}$   \\
         Qwen-VL-Chat-9B   & $1.0_{0.00}$ & $83.1_{0.02}$ & $0.9_{0.01}$ & $87.9_{0.16}$ & $2.0_{0.01}$ & $74.7_{0.12}$ & $\underline{2.9_{0.01}}$ & $\underline{68.1_{0.03}}$    & $\underline{3.0_{0.01}}$ & $\underline{74.5_{0.12}}$ & $\underline{2.9_{0.02}}$ & $\underline{75.7_{0.20}}$   \\
         IDEFICS2-8B      & $\underline{0.8_{0.01}}$ & $\underline{93.0_{0.07}}$    & $\underline{0.2_{0.00}}$ & $\underline{98.1_{0.04}}$  & $\mathbf{1.0_{0.00}}$ & $\mathbf{91.5_{0.02}}$ &  $3.0_{0.00}$ & $75.0_{0.00}$  & $5.0_{0.00}$ & $57.7_{0.08}$ & $5.0_{0.00}$ & $58.7_{0.02}$   \\
         Deepseek-VL2-Small-16B   & $4.0_{0.00}$ & $66.6_{0.00}$ & $3.7_{0.01}$ & $69.1_{0.12}$  & $3.0_{0.00}$ & $75.0_{0.00}$ & $3.0_{0.00}$ & $75.0_{0.00}$  & $5.0_{0.00}$ & $58.3_{0.00}$ & $4.9_{0.00}$ & $58.8_{0.00}$  \\ 
         OpenFlamingo-9B     & $4.0_{0.00}$ & $66.6_{0.00}$ & $4.0_{0.00}$ & $67.6_{0.00}$  & $3.0_{0.00}$ & $75.0_{0.00}$  & $3.0_{0.00}$ & $75.0_{0.00}$   & $5.0_{0.00}$ & $58.3_{0.00}$ & $5.0_{0.00}$  & $58.3_{0.00}$   \\ 
         Otter-9B & $5.0_{0.00}$ & $50.0_{0.00}$ & $4.9_{0.01}$ & $49.6_{0.12}$ & $4.0_{0.00}$ & $50.2_{0.00}$  & $5.0_{0.00}$ & $50.0_{0.03}$ & $5.2_{0.02}$ & $51.4_{0.18}$ & $3.7_{0.00}$ & $62.4_{0.20}$ \\ 
         Deepseek-VL2-27B       & $4.0_{0.0}$ & $66.7_{0.0}$ & $4.0_{0.00}$ & $66.7_{0.00}$  &  $3.0_{0.00}$ & $75.0_{0.00}$ &  $3.0_{0.00}$ & $75.0_{0.00}$    & $5.0_{0.00}$ & $58.3_{0.00}$  & $5.0_{0.00}$ & $58.3_{0.00}$   \\
         Qwen2.5-VL-Instruct-32B & $\mathbf{0.0_{0.00}}$ & $\mathbf{100.0_{0.00}}$ & $\mathbf{0.0_{0.0}}$ & $\mathbf{100.0_{0.0}}$ & $\underline{2.9_{0.01}}$ & $\underline{75.1_{0.04}}$ & $\mathbf{2.3_{0.0}}$ & $\mathbf{80.1_{0.1}}$ & $\mathbf{2.9_{0.03}}$ & $\mathbf{75.5_{0.28}}$ & $4.6_{0.0}$ & $62.1_{0.1}$ \\ 
         \midrule
         Gemini-2.0-Flash  & $\mathbf{0.0_{0.0}}$ & $\mathbf{100.0_{0.0}}$  & $0.7_{0.0}$ & $94.4_{0.0}$ & $\mathbf{1.0_{0.0}}$ & $\mathbf{91.6_{0.0}}$   & $\mathbf{2.3_{0.0}}$ & $\mathbf{80.7_{0.0}}$  & $3.2_{0.0}$ & $73.2_{0.0}$ & $\mathbf{2.8_{0.0}}$ & $\mathbf{76.9_{0.0}}$   \\
    \bottomrule
    \end{tabular}}
\end{table*}

\section{Experiments}
In this section, we empirically evaluate state-of-the-art LVLMs on our proposed tasks. Detailed data generation mechanisms for each dataset, prompt templates for each task, and LVLM model descriptions and parameters are all deferred to the Appendix.

\subsection{Datasets}
 To evaluate formal causal reasoning capabilities from visual cues, we opt to adapt and evaluate LVLMs on existing causal representation learning datasets. Our benchmark consists of three datasets, each representing a physical system, adapted to generate data satisfying each task's requirements. Since the original datasets consist of continuous-valued ground-truth factors, we discretize and convert them into textual categories. The \textbf{Pendulum} dataset~\cite{DBLP:conf/cvpr/YangLCSHW21} is a two-dimensional physical system dataset depicting a pendulum, a light source, and a shadow. The \textbf{Water Flow} dataset~\cite{DBLP:conf/cvpr/YangLCSHW21} is a two-dimensional physical system dataset depicting a red ball dropped in a glass filled with water and a hole on the right side that leaks water. The \textbf{Causal Circuit} dataset~\cite{brehmer2022weakly} is a three-dimensional physical system dataset consisting of a robot arm interacting with three colored lights. The causal variables and their relationships for each dataset are shown in~\cref{fig:examples_datasets}.

\subsection{Setup}
\paragraph{Models.} We evaluate a wide range of open-source LVLMs on the causal structure inference, intervention target prediction, and counterfactual prediction tasks, including LLaVa-Onevision (7B)~\cite{li2024llavaonevisioneasyvisualtask}, Qwen-VL-Chat (9B)~\cite{bai2023qwenvlversatilevisionlanguagemodel}, Qwen2.5-VL-Instruct (32B)~\cite{bai2025qwen2}, IDEFICS2 (8B)~\cite{laurenccon2024matters}, DeepSeek-VL2 (16B, 27B)~\cite{wu2024deepseekvl2mixtureofexpertsvisionlanguagemodels}, OpenFlamingo (9B)~\cite{awadalla2023openflamingoopensourceframeworktraining}, Otter-Llama (9B)~\cite{li2023ottermultimodalmodelincontext}, and Gemini 2.0 Flash~\cite{gemini}. We conduct our experiments on NVIDIA A100 GPUs with 40GB RAM.

\paragraph{Metrics.} For all tasks, we use the accuracy metric against the ground truth via exact match as the evaluation metric. For both the causal structure inference tasks, we ask a series of Yes/No questions for each image/pair. We construct a binary adjacency matrix according to the predicted model answers and compute the Structural Hamming Distance (SHD) with respect to the ground-truth causal graph. We also report the average exact match accuracy of model predictions. In the intervention target prediction task, we evaluate the number of times the model predicted the correct intervention target variable. For the counterfactual prediction task, we evaluate the number of times the model predicted the correct counterfactual states for each variable intervened upon. We compute the average performance over all samples in the query set across 3 random seeds for each shot. We evaluate Gemini only once due to rate limits.

\begin{table*}[htb]
    \renewcommand\arraystretch{1.0}
    \centering
    \caption{Results for \textbf{Task 2: Intervention Target Prediction} task under Zero Shot (ZS) and Few Shot (FS) settings.}
    \label{tab:intervention_results}
    \setlength{\tabcolsep}{2pt}
    \resizebox{\textwidth}{!}{ 
    \begin{tabular}{lccccccccccccc}
    \toprule
         \multirow{4}{*}{\bf{Model}} & \multicolumn{4}{c}{\textbf{Pendulum}} & \multicolumn{4}{c}{\textbf{Water Flow}} & \multicolumn{4}{c}{\textbf{Causal Circuit}} \\ 
         \cmidrule(lr){2-5} \cmidrule(lr){6-9} \cmidrule(lr){10-13}
         & \textbf{ZS} & \multicolumn{3}{c}{\textbf{FS}} 
         & \textbf{ZS} & \multicolumn{3}{c}{\textbf{FS}} 
         & \textbf{ZS} & \multicolumn{3}{c}{\textbf{FS}} \\ 
         \cmidrule(lr){3-5} \cmidrule(lr){7-9} \cmidrule(lr){11-13}
          & \textbf{0} & \textbf{2} & \textbf{4} & \textbf{8} 
         & \textbf{0} & \textbf{2} & \textbf{4} & \textbf{8} 
         & \textbf{0} & \textbf{2} & \textbf{4} & \textbf{8} \\ \midrule
         LLaVA-OneVision-7B   & $26.2_{1.5}$ & $27.5_{1.9}$ & $26.3_{1.0}$ & $\underline{27.1_{0.9}}$ & $43.1_{0.8}$ & $34.1_{2.3}$ & $34.1_{1.2}$ & $32.7_{1.2}$ & $39.4_{0.5}$ & $35.0_{0.4}$ & $\underline{36.1_{0.5}}$ & $\underline{35.9_{0.4}}$  \\
         Qwen-VL-Chat-9B  & $24.9_{0.5}$ & $24.8_{1.0}$ & $24.3_{1.4}$ & $24.7_{1.6}$ & $37.8_{0.6}$ & $33.1_{1.2}$ & $32.9_{0.8}$ & $32.1_{0.8}$ & $10.4_{0.9}$ & $31.0_{0.4}$ & $31.8_{1.6}$ & $33.0_{2.3}$ \\
         IDEFICS2-8B & $29.0_{0.4}$ & $24.2_{1.9}$ & $24.8_{0.9}$ & $24.3_{1.1}$ & $34.8_{2.1}$ & $35.4_{1.8}$ & $33.3_{0.3}$ & $33.5_{0.8}$ & $10.2_{0.4}$ & $30.3_{1.2}$ & $31.4_{0.9}$ & $29.7_{0.5}$ \\
         Deepseek-VL2-Small-16B  & $25.5_{1.1}$ & $24.4_{0.4}$ & $24.0_{0.3}$ & $0.0_{0.0}$ & $35.8_{0.6}$ & $34.4_{0.2}$ & $34.3_{0.7}$ & $0.0_{0.0}$ & $\mathbf{72.9_{1.1}}$ & $28.1_{1.5}$ & $0.2_{0.1}$ & $0.0_{0.0}$ \\ 
         OpenFlamingo-9B & $24.8_{0.5}$ & $24.7_{0.7}$ & $23.7_{1.1}$ & $25.2_{0.6}$ & $34.2_{1.7}$ & $34.5_{1.4}$ & $33.0_{1.1}$ & $33.1_{0.8}$ & $9.8_{0.6}$ & $31.6_{1.5}$ & $31.9_{2.3}$ & $32.3_{1.1}$ \\ 
         Otter-9B & $26.6_{1.9}$ & $25.3_{0.3}$ & $26.9_{0.4}$ & $23.0_{1.2}$ & $32.8_{1.1}$ & $34.1_{1.0}$ & $30.0_{0.9}$ & $31.9_{0.9}$ & $9.1_{0.7}$ & $25.2_{1.4}$ & $23.4_{1.4}$ & $24.3_{1.4}$ \\ 
         Deepseek-VL2-27B & $31.9_{0.0}$ & $\underline{30.4_{0.0}}$ & $24.1_{0.0}$ & - & $\underline{44.4_{0.0}}$ & $36.6_{0.0}$ & $31.4_{0.0}$ & - & $\underline{66.1_{0.0}}$ & $\mathbf{43.7_{0.0}}$ & $30.3_{0.0}$ & - \\ 
         Qwen2.5-VL-Instruct-32B & $\mathbf{44.3_{0.5}}$ & $29.5_{0.3}$ & $\underline{27.4_{2.0}}$ & $26.2_{1.2}$ & $\mathbf{48.4_{0.7}}$ & $\underline{37.2_{1.3}}$ & $\underline{37.3_{0.7}}$ & $\underline{36.6_{0.6}}$ & $32.1_{1.5}$ & $32.5_{0.9}$ & $32.0_{1.2}$ & $34.6_{0.8}$ \\ 
         \midrule
         Gemini-2.0-Flash   & $\underline{39.4_{0.0}}$ & $\mathbf{45.2_{0.0}}$  & $\mathbf{45.3_{0.0}}$  & $\mathbf{47.4_{0.0}}$  & $37.6_{0.0}$  & $\mathbf{46.5_{0.0}}$  & $\mathbf{52.4_{0.0}}$ & $\mathbf{55.7_{0.0}}$   & $10.5_{0.0}$ & $\underline{43.1_{0.0}}$ & $\mathbf{55.1_{0.0}}$  & $\mathbf{66.1_{0.0}}$ \\
    \bottomrule
    \end{tabular}}
\end{table*}

\begin{table*}[ht]
    \renewcommand\arraystretch{1.0}
    \centering
    \caption{Results for \textbf{Task 3: Counterfactual Prediction} task under Zero Shot (ZS) and Few Shot (FS) settings.}
    \label{tab:counterfactual_results}
    \setlength{\tabcolsep}{2pt}
    \resizebox{\textwidth}{!}{
    \begin{tabular}{lccccccccccccc}
    \toprule
         \multirow{4}{*}{\bf{Model}} & \multicolumn{4}{c}{\textbf{Pendulum}} & \multicolumn{4}{c}{\textbf{Water Flow}} & \multicolumn{4}{c}{\textbf{Causal Circuit}} \\ 
         \cmidrule(lr){2-5} \cmidrule(lr){6-9} \cmidrule(lr){10-13}
         & \textbf{ZS} & \multicolumn{3}{c}{\textbf{FS}} 
         & \textbf{ZS} & \multicolumn{3}{c}{\textbf{FS}} 
         & \textbf{ZS} & \multicolumn{3}{c}{\textbf{FS}} \\ 
         \cmidrule(lr){3-5} \cmidrule(lr){7-9} \cmidrule(lr){11-13}
         & \textbf{0} & \textbf{2} & \textbf{4} & \textbf{8} 
         & \textbf{0} & \textbf{2} & \textbf{4} & \textbf{8} 
         & \textbf{0} & \textbf{2} & \textbf{4} & \textbf{8} \\ \midrule
         LLaVA-OneVision-7B  & $\mathbf{84.1_{0.8}}$ & $83.0_{0.7}$ & $83.6_{0.5}$ & $83.5_{0.5}$ & $\underline{83.4_{0.4}}$ & $\underline{84.0_{0.3}}$ & $\underline{84.5_{0.7}}$ & $85.0_{0.6}$ & $94.9_{0.2}$ & $96.0_{0.2}$ & $96.7_{0.2}$ & $96.9_{0.2}$ \\
         Qwen-VL-Chat-9B & $80.1_{0.9}$ & $81.5_{0.6}$ & $80.1_{1.1}$ & $73.4_{1.1}$ & $\mathbf{83.4_{0.3}}$ & $82.3_{0.2}$ & $82.6_{0.4}$ & $81.2_{0.3}$ & $74.4_{0.3}$ & $94.7_{0.1}$ & $94.4_{0.0}$ & $94.9_{0.1}$ \\
         IDEFICS2-8B & $38.8_{0.5}$ & $79.3_{0.6}$ & $80.2_{0.4}$ & $80.3_{0.5}$ & $71.0_{0.3}$ & $82.7_{0.7}$ & $83.7_{0.3}$ & $84.1_{0.3}$ & $58.1_{0.3}$ & $92.2_{0.5}$ & $95.4_{0.3}$ & $96.8_{0.1}$ \\
         Deepseek-VL2-Small-16B  & $77.9_{0.6}$ & $20.4_{0.4}$ & $0.0_{0.0}$ & $0.0_{0.0}$ & $51.4_{0.3}$ & $71.3_{0.6}$ & $2.6_{0.4}$ & $0.0_{0.0}$ & $41.6_{1.0}$ & $64.9_{0.4}$ & $21.8_{0.4}$ & $0.0_{0.0}$ \\ 
         OpenFlamingo-9B & $22.4_{1.1}$ & $82.6_{0.8}$ & $81.8_{0.8}$ & $81.7_{0.8}$ & $33.0_{1.1}$ & $80.6_{0.6}$ & $81.1_{0.7}$ & $83.9_{0.5}$ & $3.9_{0.3}$ & $87.9_{0.1}$ & $93.2_{0.3}$ & $88.9_{0.3}$  \\ 
         Otter-9B & $22.6_{0.5}$ & $13.4_{0.3}$ & $1.7_{0.2}$ & $0.4_{0.1}$ & $34.0_{0.3}$ & $32.0_{0.6}$ & $25.4_{0.3}$ & $6.2_{0.2}$ & $43.8_{0.3}$ & $53.3_{0.4}$ & $37.7_{0.1}$ & $6.3_{0.2}$ \\ 
         Deepseek-VL2-27B & $61.1_{0.0}$ & $83.5_{0.0}$ & $83.7_{0.0}$ & - & $83.0_{0.0}$ & $83.0_{0.0}$ & $83.2_{0.0}$ & - & $94.2_{0.0}$ & $94.0_{0.0}$ & $94.8_{0.0}$ & - \\ 
         Qwen2.5-VL-Instruct-32B & $81.8_{0.5}$ & $\mathbf{85.1_{0.5}}$ & $\mathbf{85.5_{0.8}}$ & $\mathbf{87.4_{0.8}}$ & $79.9_{0.4}$ & $82.6_{0.1}$ & $84.4_{0.4}$ & $\underline{86.7_{0.7}}$ & $\mathbf{99.1_{0.1}}$ & $\mathbf{98.8_{0.1}}$ & $\mathbf{98.6_{0.1}}$ & $\mathbf{98.4_{0.0}}$ \\
         \midrule
         Gemini-2.0-Flash   & $\underline{83.4_{0.0}}$ & $\underline{84.9_{0.0}}$  & $\underline{85.0_{0.0}}$   & $\underline{86.5_{0.0}}$  & $80.3_{0.0}$  & $\mathbf{84.3_{0.0}}$   & $\mathbf{86.5_{0.0}}$ & $\mathbf{88.3_{0.0}}$   & $\underline{97.0_{0.0}}$ & $\underline{97.4_{0.0}}$ & $\underline{97.2_{0.0}}$  & $\underline{97.4_{0.0}}$    \\
    \bottomrule
    \end{tabular}}
\end{table*}

\subsection{Results}
We include the results for Task 1A and 1B in~\cref{tab:causal_structure_results}, Task 2 in~\cref{tab:intervention_results}, and Task 3 in~\cref{tab:counterfactual_results} for all three datasets using all models, where the best performance is bold, and the second-best is underlined in each column. We observe that in-context learning for causal reasoning tasks is only marginally effective and, in some cases, can degrade model performance for the majority of open-source models. Most models degrade in performance with increasing shots. Qwen2.5-VL, which is much larger, is an exception to this trend and notably improves with in-context demonstration examples, but for only the counterfactual prediction task. Gemini-2.0-Flash also has the same upward trend for the few-shot setting. However, all LVLMs struggle in tasks requiring multi-image inputs, such as the intervention target prediction task. Although some models, such as LLaVA-OneVision, were specifically trained with interleaved multi-image inputs, they demonstrate subpar performance in complex reasoning scenarios. In the following, we analyze our experimental results for each task in more detail, weighing the pros and cons of different models and observing general trends.

\paragraph{Causal Structure Inference Results.} For the causal structure inference tasks, both standard and interleaved shown in~\cref{tab:causal_structure_results}, the best performing models are Qwen2.5-VL, Gemini-2.0-Flash across all datasets. However, IDEFICS2 and Qwen-VL-Chat show comparable performance for Pendulum and Flow. LLaVA-OneVision-7B is generally the most consistent small open-source model across all datasets, with its performance bounded between other smaller and larger models. We also observe that performance degrades for most models when provided with paired images for causal structure inference. Among all models, Qwen2.5-VL and Gemini predict the true causal graph for the Pendulum dataset with perfect accuracy. We conjecture this is due to the simple nature of the causal mechanism in the Pendulum dataset. However, it is not clear how this ability scales as the causal graph increases in size. Our results also indicate that LVLMs predict the causal graphs for the pendulum and water flow datasets more accurately compared to the causal circuit dataset. This is most likely because, unlike the pendulum and water flow systems, the causal graph in the causal circuit system is induced and not a naturally occurring phenomenon. In such cases, providing a pair of images through the interleaved variant often improves the inference performance. A more nuanced analysis of the causal structure inference task, including precision and recall results to gauge how frequently models default to ``Yes'' or ``No'' answers, and evaluating whether models can distinguish between directionality and cyclicity, is provided in Appendix~\ref{app:additional_results}. 

\begin{table*}
    \renewcommand\arraystretch{1.0}
    \small
    \centering
    \caption{Selected results for \textbf{Task 2: Intervention Target Prediction} task under Zero Shot (ZS) and Few Shot (FS) settings \textbf{without causal graph}.}
    \label{tab:intervention_nograph_results}
    \setlength{\tabcolsep}{2pt}
    \resizebox{\textwidth}{!}{
    \begin{tabular}{lccccccccccccc}
    \toprule
         \multirow{4}{*}{\bf{Model}} & \multicolumn{4}{c}{\textbf{Pendulum}} & \multicolumn{4}{c}{\textbf{Water Flow}} & \multicolumn{4}{c}{\textbf{Causal Circuit}} \\ 
         \cmidrule(lr){2-5} \cmidrule(lr){6-9} \cmidrule(lr){10-13}
         & \textbf{ZS} & \multicolumn{3}{c}{\textbf{FS}} 
         & \textbf{ZS} & \multicolumn{3}{c}{\textbf{FS}} 
         & \textbf{ZS} & \multicolumn{3}{c}{\textbf{FS}} \\ 
         \cmidrule(lr){3-5} \cmidrule(lr){7-9} \cmidrule(lr){11-13}
          & \textbf{0} & \textbf{2} & \textbf{4} & \textbf{8} 
         & \textbf{0} & \textbf{2} & \textbf{4} & \textbf{8} 
         & \textbf{0} & \textbf{2} & \textbf{4} & \textbf{8} \\ \midrule
         LLaVA-OneVision-7B   & $26.2_{1.5}$ & $26.7_{1.8}$ & $26.2_{0.5}$ & $26.4_{0.7}$ & $39.6_{1.1}$ & $34.0_{2.2}$ & $34.0_{0.7}$ & $32.6_{1.5}$ & $39.3_{1.2}$ & $35.2_{0.1}$ & $36.2_{0.3}$ & $35.8_{0.5}$    \\
         Qwen2.5-VL-Instruct-32B & $44.4_{2.4}$ & $34.8_{2.4}$ & $29.8_{1.5}$ & $26.8_{1.3}$ & $54.6_{2.0}$ & $39.2_{2.9}$ & $35.4_{0.4}$ & $31.2_{1.4}$ & $18.8_{1.1}$ & $28.0_{0.7}$ & $30.8_{0.3}$ & $38.2_{0.9}$   \\ 
         Gemini-Flash-2.0 & $35.6_{0.0}$ & $46.4_{0.0}$ & $49.5_{0.0}$ & $48.2_{0.0}$ & $37.4_{0.0}$ & $46.7_{0.0}$ & $50.6_{0.0}$ & $53.6_{0.0}$ & $18.8_{0.0}$ & $75.8_{0.0}$ & $81.8_{0.0}$ & $81.4_{0.0}$ \\
    \bottomrule
    \end{tabular}}
\end{table*}

\begin{table*}
    \renewcommand\arraystretch{1.0}
    \centering
    \small
    \caption{Selected results for the \textbf{Task 3: Counterfactual Prediction} task under Zero Shot (ZS) and Few Shot (FS) settings \textbf{without causal graph}.}
    \label{tab:counterfactual_nograph_results}
    \setlength{\tabcolsep}{2pt}
    \resizebox{\textwidth}{!}{
    \begin{tabular}{lccccccccccccc}
    \toprule
         \multirow{4}{*}{\bf{Model}} & \multicolumn{4}{c}{\textbf{Pendulum}} & \multicolumn{4}{c}{\textbf{Water Flow}} & \multicolumn{4}{c}{\textbf{Causal Circuit}} \\ 
         \cmidrule(lr){2-5} \cmidrule(lr){6-9} \cmidrule(lr){10-13}
         & \textbf{ZS} & \multicolumn{3}{c}{\textbf{FS}} 
         & \textbf{ZS} & \multicolumn{3}{c}{\textbf{FS}} 
         & \textbf{ZS} & \multicolumn{3}{c}{\textbf{FS}} \\ 
         \cmidrule(lr){3-5} \cmidrule(lr){7-9} \cmidrule(lr){11-13}
         & \textbf{0} & \textbf{2} & \textbf{4} & \textbf{8} 
         & \textbf{0} & \textbf{2} & \textbf{4} & \textbf{8} 
         & \textbf{0} & \textbf{2} & \textbf{4} & \textbf{8} \\ \midrule
         LLaVA-OneVision-7B  & $83.5_{0.8}$ & $83.0_{0.9}$ & $83.3_{0.7}$ & $83.2_{0.7}$ & $82.4_{0.3}$ & $83.5_{0.2}$ & $83.7_{0.5}$ & $84.8_{0.2}$ & $94.0_{0.1}$ & $95.9_{0.1}$ & $96.4_{0.2}$ & $96.7_{0.2}$    \\
         Qwen2.5-VL-Instruct-32B & $79.0_{0.4}$ & $83.7_{0.8}$ & $85.2_{0.7}$ & $87.5_{0.8}$ & $82.2_{0.6}$ & $84.4_{0.2}$ & $85.3_{0.5}$ & $86.9_{0.8}$ & $94.9_{0.1}$ & $95.4_{0.1}$ & $96.5_{0.1}$ & $97.3_{0.0}$ \\
         Gemini-Flash-2.0 & $74.8_{0.0}$ & $81.5_{0.0}$ & $83.2_{0.0}$ & $86.9_{0.0}$ & $76.6_{0.0}$ & $83.6_{0.0}$ & $85.3_{0.0}$ & $88.0_{0.0}$  & $82.9_{0.0}$ & $93.0_{0.0}$ & $93.6_{0.0}$ & $94.6_{0.0}$ \\
    \bottomrule
    \end{tabular}}
\end{table*}

\paragraph{Intervention Target Prediction Results.} The intervention target prediction task is progressively more difficult. Conditioned on the given causal structure, the model is required to reason about the intervened variable that caused the change between the first and second image. In~\cref{tab:intervention_results}, the best-performing models are DeepseekVL2, Qwen2.5-VL, and Gemini-2.0-Flash. Among these models, Gemini demonstrates the best trend in improving intervention target prediction with an increasing number of shots. For the causal circuit dataset, originating from an induced causal graph, DeepseekVL2-Small and DeepSeekVL2 have notable zero-shot performance. We attribute this to the reasoning-focused training paradigm of DeepSeekVL2, which is better suited for reasoning with a concrete set of rules. However, for open-source models, providing few-shot examples generally degrades model performance. Similar to the causal structure inference task, the most consistently performing model is LLaVA-OneVision-7B. 
Looking more closely at model predictions, we observe that several models tend to predict one variable more than others during zero-shot inference. For instance, in the pendulum dataset, some models predict \textit{light position} for most test queries. We conjecture this is due to the light being the most noticeable object in the image. Note that the frequently predicted variable may vary across models.

\paragraph{Counterfactual Prediction Results.} The counterfactual prediction task involves predicting exact discretized values of each causal variable had a given intervention occurred. Ideally, interventions on variables should propagate to accurate values for descendants and should not change the values for ancestors. We evaluate this task across different levels of granularity. First, we compute model correctness as the average accuracy over all variables. However, this metric can easily be inflated as the model can achieve some level of correctness by simply predicting the initial input states. As a result, we observe seemingly favorable performance in~\cref{tab:counterfactual_results}. Therefore, to understand how interventions on each variable affect utility, we show the per-variable breakdown for each dataset in Appendix~\ref{app:additional_results}. We observe that most models attain better results when the intervention is performed on variables with no descendants, but struggle with accurately propagating causal changes to descendants. Generally, LLaVA-OneVision-7B, Deepseek-VL2, Qwen2.5-VL, and Gemini-2.0-Flash have the best performance. We also look at interventions on variables with at least one descendant and compute the average accuracy of the correct prediction of downstream variables. We find that most models achieve notably lower performance as they are not able to correctly predict propagated descendant values. Nonetheless, most models improve in their ability to correctly predict descendant variable values as we increase the number of shots.

\begin{table*}[htb]
    \renewcommand\arraystretch{1.0}
    \centering
    \small
    \caption{Selected results for \textbf{Task 2: Intervention Target Prediction} under \textbf{Balanced Selection} \textbf{with causal graph}.}
    \label{tab:intervention_balanced}
    \setlength{\tabcolsep}{4pt}
    \begin{tabular}{lcccccc}
    \toprule
         \multirow{2}{*}{\bf{Model}} & \multicolumn{2}{c}{\textbf{Pendulum}} & \multicolumn{2}{c}{\textbf{Water Flow}} & \multicolumn{2}{c}{\textbf{Causal Circuit}} \\ 
         \cmidrule(lr){2-3} \cmidrule(lr){4-5} \cmidrule(lr){6-7}
         & \textbf{4-shot} & \textbf{8-shot}
         & \textbf{3-shot} & \textbf{6-shot}
         & \textbf{4-shot} & \textbf{8-shot} \\ \midrule
         LLaVA-OneVision-7B   & $26.5_{0.0}$ & $26.5_{0.0}$  & $31.6_{0.0}$ & $31.2_{0.0}$  & $36.4_{0.0}$ & $36.4_{0.0}$  \\
         Qwen2.5-VL-Instruct-32B       & $32.5_{0.0}$ & $25.3_{0.0}$  & $45.3_{0.0}$  & $36.5_{0.0}$  & $40.4_{0.0}$ & $14.8_{0.0}$  \\ 
         Gemini-Flash-2.0     & $46.2_{0.0}$ & $44.3_{0.0}$  & $43.2_{0.0}$ & $55.3_{0.0}$  & $48.1_{0.0}$ & $41.6_{0.0}$  \\
    \bottomrule
    \end{tabular}
\end{table*}

\begin{table*}[htb]
    \renewcommand\arraystretch{1.0}
    \centering
    \small
    \caption{Selected results for all tasks under \textbf{Zero-Shot Chain of Thought (CoT)} setting.}
    \label{tab:cot}
    \setlength{\tabcolsep}{2pt}
    \resizebox{\textwidth}{!}{
    \begin{tabular}{lccccccccccccc}
    \toprule
         \multirow{3}{*}{\bf{Model}} & \multicolumn{4}{c}{\textbf{Pendulum}} & \multicolumn{4}{c}{\textbf{Water Flow}} & \multicolumn{4}{c}{\textbf{Causal Circuit}} \\ 
         \cmidrule(lr){2-5} \cmidrule(lr){6-9} 
         \cmidrule(lr){3-5} \cmidrule(lr){7-9} \cmidrule(lr){10-13}
         & \textbf{1A} & \textbf{1B} & \textbf{2} & \textbf{3} 
         & \textbf{1A} & \textbf{1B} & \textbf{2} & \textbf{3}
         & \textbf{1A} & \textbf{1B} & \textbf{2} & \textbf{3} \\ \midrule
         Qwen2.5-VL-Instruct-32B & $98.7_{0.3}$ & $87.4_{0.2}$ & $36.5_{0.0}$ & $81.6_{0.6}$ & $84.0_{0.1}$ & $74.9_{0.6}$ & $33.0_{0.0}$ & $80.6_{0.2}$ & $66.3_{0.4}$ & $59.2_{0.0}$ & $54.8_{0.0}$ & $98.5_{0.2}$ \\
         Gemini-Flash-2.0 & $98.1_{0.0}$ & $87.5_{0.0}$ & $39.7_{0.0}$ & $83.7_{0.0}$ & $92.3_{0.0}$ & $83.8_{0.0}$ & $40.4_{0.0}$ & $80.4_{0.0}$ & $70.7_{0.0}$ & $72.4_{0.0}$ & $9.1_{0.0}$ & $96.1_{0.0}$ \\
    \bottomrule
    \end{tabular}}
\end{table*}

\subsection{Additional Analysis}
\paragraph{Inference without Causal Graph.} We investigate the ability of LVLMs to reason about causal concepts when ground-truth causal relationships are not provided in the input. We evaluate the performance of the top-performing models for Tasks 2 and 3. The results in~\cref{tab:intervention_nograph_results} and~\cref{tab:counterfactual_nograph_results} demonstrate that model performance generally degrades, albeit marginally in some cases, when the causal relationships are not provided. We note that the Pendulum and Flow datasets follow natural physical laws, whereas the Causal Circuit dataset has an induced causal graph. If the causal system is derived from physical laws, the model is more likely to be familiar with the governing causal mechanism. In datasets with induced causal graphs, it may be beneficial to provide the causal relationships as additional context. However, contrary to intuition, we find that excluding the causal structure information significantly improves the intervention target prediction performance of Gemini-2.0-Flash on the causal circuit dataset.

\paragraph{Balanced Demonstration Selection.} In this experiment, we evaluate the influence of the demonstration selection strategy on model predictions by implementing a balanced selection technique compared to the random selection used in prior experiments. We assess the method on the intervention target prediction task as it is flexible enough to incorporate the strategy. Specifically, we construct a demonstration set such that each causal variable appears as an intervention target. The balanced demonstrations ensure that the model can contextualize the influence of all possible interventions on the given system. The results from this experiment are shown in~\cref{tab:intervention_balanced}. Compared to the random selection strategy, we observe that providing intervention examples representative of each causal variable does not improve prediction performance. Also, some open-source models, such as Qwen2.5, rely heavily on the first demonstration example. Our findings demonstrate that open-source models still struggle to make reliable inferences from few-shot examples in reasoning tasks.

\paragraph{Chain-of-Thought Prompting (CoT).}
A common technique to improve the reasoning ability of LLMs/LVLMs is chain-of-thought (CoT) prompting~\cite{wei2022chain, kojima2022large}. Here we study how chain-of-thought prompting affects predictive accuracy in zero-shot inference. We construct a CoT prompt for each task as follows: We prompt the model to obtain a reasoning chain about the objects in the images(s). Then, we augment the input with the generated response and prompt the model again for the final prediction. The results of this study are shown in~\cref{tab:cot}. Evidently, zero-shot CoT prompting improves model performance in some cases. For instance, the intervention target prediction accuracy of the Qwen2.5-VL increases significantly on the Causal Circuit dataset. Similarly, CoT improves the performance of Gemini on the Pendulum and Flow datasets. However, Gemini underperforms on the Causal Circuit dataset with CoT. Possibly, adding a lengthy reasoning chain to the input takes up the model context length, negatively affecting predictive performance. Based on these observations, it is inconclusive whether zero-shot chain-of-thought prompting can truly improve causal reasoning in open-source models.

\section{Conclusion}
In this paper, we introduce \textit{CausalVLBench}, a benchmark for evaluating the visual causal reasoning capability of large vision-language models. We formulate three different tasks: causal structure inference, intervention target prediction, and counterfactual prediction, and rigorously evaluate the performance of open-source LVLMs under zero-shot and few-shot settings. Furthermore, we observe the effect of removing causal relationships from the prompt, chain-of-thought prompting, and balanced demonstration selection on model performance. Our results indicate that larger models with more parameters are necessary for effective visual causal reasoning from in-context learning, and smaller open-source models exhibit poor performance and still need further optimization for few-shot learning. Furthermore, our observations serve as motivation to consider causal reasoning as a core element of new training paradigms.

\section*{Acknowledgements}
This work is supported in part by National Science Foundation under awards 1910284, 1946391, 2119691, 2147375, the National Institute of General Medical Sciences of National Institutes of Health under award P20GM139768, and the Arkansas Integrative Metabolic Research Center at University of Arkansas.

\section*{Limitations}
In this work, we construct a comprehensive benchmark to evaluate visual causal reasoning capabilities of large vision-language models. To ensure reproducibility and transparency, we run our evaluations primarily on open-weight LVLMs and include one closed-source model. Due to resource constraints, we do not evaluate on closed models such as GPT and Claude and encourage future evaluations with these proprietary models.

Our benchmark is largely designed to gauge the causal reasoning abilities of existing LVLMs on common datasets from causal representation learning. We focus on systems with four-variable causal graphs for the purposes of our work. We observe that LVLMs struggle on even simple causal graphs and leave evaluations on more complex causal systems, including temporal systems, for future work.

\section*{Ethical Considerations}
 Due to the increasing use of large foundation models in real-world applications, it is paramount to study their reasoning capabilities to develop more robust and reliable AI systems. In this work, we focus on benchmarking large vision-language models on formal causal reasoning. The datasets used in this paper are publicly available. Our use of publicly available LVLMs and the proprietary Gemini-2.0-Flash adhere to the associated licenses. We hope that our benchmark inspires the development of new training paradigms with an emphasis on causal reasoning.

\bibliography{main}

\appendix

\section*{Appendices}

\section{Implementation Details}
\label{app:implementation}

\subsection{Model Configurations}
We briefly discuss each model benchmarked in this evaluation below and provide configuration details for all models in~\cref{tab:models}, including the number of image tokens, context length, and architecture.

\paragraph{OpenFlamingo-9B.} OpenFlamingo-9B~\cite{awadalla2023openflamingoopensourceframeworktraining} is an auto-regressive open-source vision-language model that allows the mapping of interleaved images and text to textual outputs. The 9B version of the model uses a CLIP ViT/14 as the vision encoder, a Perceiver resampler to extract patch features, and MPT-7B as the language model with a cross-attention module every 4th layer. The pre-training data includes the LAION-2B and MMC4 datasets, along with synthetic image-text sequences generated by ChatGPT.   

\paragraph{Otter-LLaMA-9B.} Otter-LLaMA-9B~\cite{li2023ottermultimodalmodelincontext} is an in-context instruction-tuned vision-language model based on OpenFlamingo's implementation. This framework uses an OpenFlamingo base model with CLIP ViT/14 as the vision encoder and LLaMA-7B as the language model. Otter VLM is obtained by fine-tuning the Perceiver resampler module, cross-attention layers, and input/output embeddings of a pre-trained OpenFlamingo model on the MIMIC-IT dataset. The model and its pre-trained weights are publicly available. 

\paragraph{Qwen-VL-Chat-9B.} Qwen-VL-Chat-9B~\cite{bai2023qwenvlversatilevisionlanguagemodel} is an open-source instruction fine-tuned vision-language model that adds multi-modal and multilingual capabilities to Qwen-7B language model. The model uses CLIP ViT-bigG/14 as the vision encoder, and implements a single-layer cross-attention in the vision-language adapter. Training includes three stages of learning from image-text sequences, interleaved and multi-task data, and instruction tuning data, respectively. Pre-training and fine-tuning data comprise LAION, COCO, GQA, VGQA, GRIT, and Visual Genome, among others.       

\paragraph{IDEFICS2-8B.} IDEFICS2-8B~\cite{laurenccon2024matters} is a fully auto-regressive vision-language model that incorporates SigLIP-SO400M as the vision component, Mistral-7B-v0.1 as the language module and a projection+pooling module to obtain visual tokens. The model is trained over three kinds of data, including image-text pairs from LAION COCO, interleaved image-text documents from OBELICS, and PDF documents from m OCR-IDL, PDFA, and RenderedText. The model and its pre-trained weights are publicly available.    

\paragraph{LLaVA-OneVision-7B.} LLaVA-OneVision-7B~\citep{li2024llavaonevisioneasyvisualtask} is an open, large multi-modal model trained for single-image, multi-image, and video tasks. The model architecture consists of Qwen-2 and SigLIP with a 2-layer MLP as the projector to transform image features into the text embedding space. LLaVA-OneVision is trained through three phases, namely language-image alignment, high-quality knowledge learning, and visual instruction tuning on single-image, multi-image, and video datasets.  

\paragraph{DeepseekVL2.} DeepseekVL2~\citep{wu2024deepseekvl2mixtureofexpertsvisionlanguagemodels} is a collection of open-source vision-language models which employs a mixture-of-experts model for pre-training. Both the 16B and 27B models employ SigLIP-SO400M-384 as the vision encoder, DeepSeekMoE as the mixture-of-experts language model, and a two-layer MLP as the vision-language adaptor for projecting visual tokens into the language model's embedding space. The 27B model additionally incorporates an expert bias correction step with a global bias term. The models are pre-trained on interleaved image-text, image captioning, optical character recognition, VQA, visual grounding, and grounded conversation datasets. The models are additionally fine-tuned on Supervised Fine-tuning (SFT) QA pairs encompassing alignment, understanding, reasoning, logic, conversation, and code generation. 

\paragraph{Qwen2.5-VL-Instruct-32B.} Qwen2.5-VL-Instruct-32B~\cite{bai2025qwen2} integrates Qwen 2.5 LLM as the language model and a redesigned ViT as the vision encoder to allow multi-modal input processing, including images and videos, and multi-step reasoning. The architecture also includes an MLP-based vision-language merger to obtain spatially adjacent patch features. The pre-training data is composed of multi-modal data such as image captions, optical character recognition, visual knowledge, academic questions, image/video localization, document parsing, agent-based interaction, and interleaved image-text pairs, and the post-training alignment is performed with single-turn and multi-turn instruction fine-tuning data. Model weights are publicly available.    

\paragraph{Gemini-2.0-Flash.} Gemini-2.0-Flash~\cite{gemini} is a closed-source vision-language model incorporating cross-modal attention layers to facilitate input processing from multiple modalities. The model is pre-trained on text corpora including books, research articles, and text crawled from the web, image-text pairs, temporal datasets, and multilingual audio datasets. The model is further fine-tuned on domain-specific datasets. Details about model architectures and datasets used are not publicly available.

\section{Data Generation}
\label{app:data_gen}

\subsection{Dataset Descriptions}
\paragraph{Pendulum.} The Pendulum dataset~\cite{DBLP:conf/cvpr/YangLCSHW21} is a synthetic dataset that consists of $7$K images with resolution $96\times 96\times 4$ generated by $4$ ground-truth causal variables: $u_1=$ pendulum angle, $u_2=$ light position, $u_3=$ shadow length, and $u_4=$ shadow position, which are continuous values. Each causal variable is determined from the following process with nonlinear functions. 

\begin{equation*}
    u_1 \sim U(-45, 45);\; \qquad \theta = u_1 * \frac{\pi}{200}
\end{equation*}

\begin{equation*}
    u_2 \sim U(60, 145);\; \qquad \phi = u_2 * \frac{\pi}{200}
\end{equation*}

\begin{align*}
    u_3 &= \max(3, \Big|9.5\frac{\cos\theta}{\tan\phi} + 9.5\sin\theta\Big|)
\end{align*}

\begin{align*}
    u_4 &= \frac{-11 + 4.75\cos\theta}{\tan\phi} + (10 + 4.75\sin\theta)
\end{align*}

\noindent
\paragraph{Water Flow.} The Flow dataset~\cite{DBLP:conf/cvpr/YangLCSHW21} is a synthetic dataset that consists of $8$K images with resolution $96\times 96\times 4$ generated by $4$ ground-truth causal variables: $u_1=$ ball radius, $u_2=$ hole position, $u_3=$ water level, and $u_4=$ water flow, which are continuous values. Each causal variable is determined by the following nonlinear physical mechanisms.
\begin{align*}
    u_1 &= \frac{r}{30}, \qquad r\in \{5, 6, \dots, 34\} \\
    u_2 &= \frac{\text{hole}}{3}, \qquad \text{hole} \in \{6, 7, \dots, 14\} \\
    u_3 &= u_1^3 + \frac{h_\text{raw}}{10}, \qquad h_{\text{raw}}\in \{10, 11, \dots, 39\} \\
    u_4 &= \sqrt{2\cdot 0.98\cdot h_w \cdot (u_3 - 0.5)}
\end{align*}

\noindent
\paragraph{Causal Circuit.} The Causal Circuit dataset is a new dataset created by~\cite{brehmer2022weakly} to explore research in causal representation learning. The dataset consists of $512\times 512\times 3$ resolution images generated by $4$ ground-truth latent causal variables: robot arm position, red light intensity, green light intensity, and blue light intensity. The images show a robot arm interacting with a system of buttons and lights. The data is rendered using an open-source physics engine. The original dataset consists of pairs of images before and after an intervention has taken place. The data is generated according to the following process:
\begin{align*}
    v_R &= 0.2 + 0.6 * \text{clip}(u_2 + u_3 + b_R, 0, 1) \\
    v_G &= 0.2 + 0.6 * b_G \\
    v_B &= 0.2 + 0.6 * b_B \\
    u_4 &\sim \text{Beta}(5v_R, 5 * (1 - v_R)) \\
    u_3 &\sim \text{Beta}(5v_G, 5 * (1 - v_G)) \\
    u_2 &\sim \text{Beta}(5v_B, 5 * (1 - v_B)) \\
    u_1 &\sim U(0, 1)
\end{align*}
where $b_R$, $b_G$, and $b_B$ are the pressed state of buttons that depends on how far the button is touched from the center, $u_1$ is the robot arm position, and $u_2$, $u_3$, and $u_4$ are the intensities of the blue, green, and red lights, respectively. From this generative process, we selectively choose only images for which the causal graph is satisfied (the robot arm's position and the downstream effects). For example, the robot arm appearing over the green button, green button lit up, and red button lit up is consistent with the assumption that the robot arm position causes changes in which buttons light up according to the causal graph. The filtered dataset consists of roughly $5$K samples.

\subsection{Dataset Construction}
Utilizing these original datasets as a base, we construct new datasets for our LVLM tasks. Specifically, for the interleaved tasks, we generate image pairs of pendulum and flow systems before and after an intervention. For the intervention target prediction task, our dataset consists of image pairs and a question about which variable was the intervention target. For the counterfactual task, we discretize the continuous numerical values of each variable into a text description and embed them into a natural language prompt. We split each dataset into $40\%$ support set and $60\%$ query set. Of the query set, we randomly sample $1000$ examples for each run during inference.

\section{Additional Results}
\label{app:additional_results}

\paragraph{Precision/Recall for Causal Structure Inference}
For a more nuanced analysis, we report precision and recall scores on the causal structure inference tasks in~\cref{tab:prec_rec_csi}. We observe that smaller models, such as Llava-OneVision have lower precision scores suggesting a tendency to stick with the "Yes" answer. Consistent with our main results, we also observe that these models perform significantly worse on precision/recall for the Causal Circuit dataset which consists of a more unintuitive causal structure. Furthermore, specifically for the Causal Circuit dataset, models such as IDEFICS (and Qwen2.5-VL for Interleaved) seem to answer "No" at a much higher rate than other models, as shown by the low recall scores. Generally, Gemini-2.0-Flash outperforms other models on precision and recall, suggesting its robustness across steps.

\paragraph{Directionality vs. Cyclicity.} 
 For the causal structure inference task, we additionally evaluate whether LVLMs can distinguish unidirectionality and acyclicity, shown in~\cref{tab:bvc}. For instance, $A \rightarrow B \rightarrow C \rightarrow A$ satisfies unidirectionality by violates acyclicity. On the other hand, all acyclic graphs are by definition unidirectional. To investigate this phenomenon, we compute two metrics, bidirectionality ($b$) and cyclicity ($c$) as follows:
\begin{equation}
    b = \frac{\text{number of bidirectional predictions}}{\text{k}}
\end{equation}
\begin{equation}
    c = \text{Tr}(e^{A\circ A}) - n
\end{equation}
where $k$ is the total number of unique variable pairs such that (X, Y) and (Y, X) are counted as the same, $A$ is the adjacency matrix, 
 $n$ is the number of nodes, and the cyclicity score $c$ implies $A$
 is a directed acyclic graph (DAG) and $c > 0$ indicates cycles~\cite{zheng2018dags}. From our results in the table below, we find that there are many more bidirectional and cyclic predictions for some models than others and that models predicting more edges as bidirectional also predict cycles at a higher rate. This indicates that the models do not distinguish between directionality and acyclicity. 

\begin{figure*}
    \centering

    \begin{subfigure}{\linewidth}
        \centering
        \includegraphics[width=0.9\linewidth]{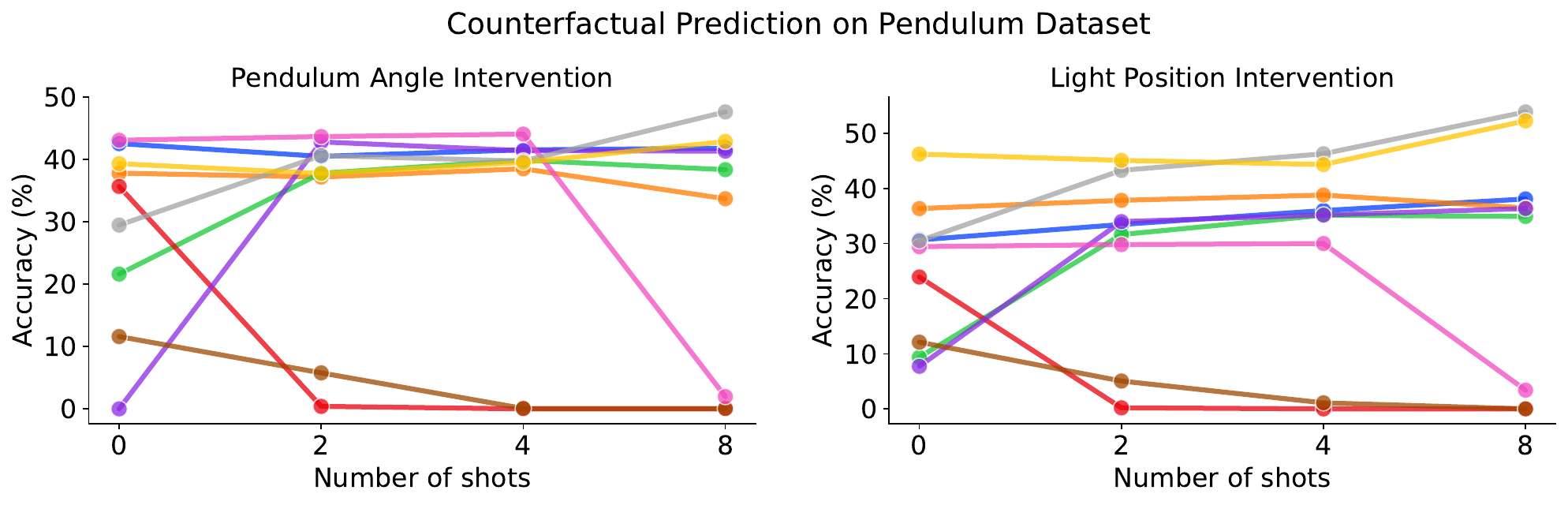}
        \caption{Average Counterfactual Prediction accuracy for pendulum interventions with at least one descendant.}
        \label{fig:cf_pendulum_plot}
    \end{subfigure}
    
    \vspace{1.0em} 

    \begin{subfigure}{\linewidth}
        \centering
        \includegraphics[width=0.9\linewidth]{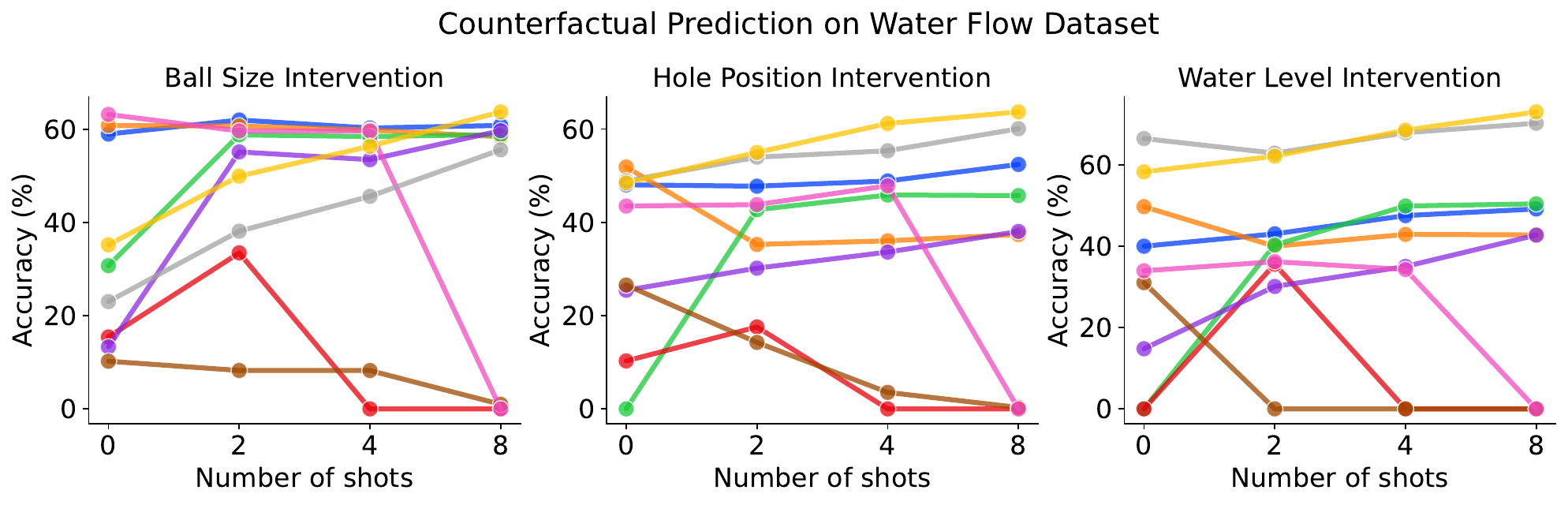}
        \caption{Average Counterfactual Prediction accuracy for water flow interventions with at least one descendant.}
        \label{fig:cf_flow_plot}
    \end{subfigure}

    \vspace{1.0em}

    \begin{subfigure}{\linewidth}
        \centering
        \includegraphics[width=0.9\linewidth]{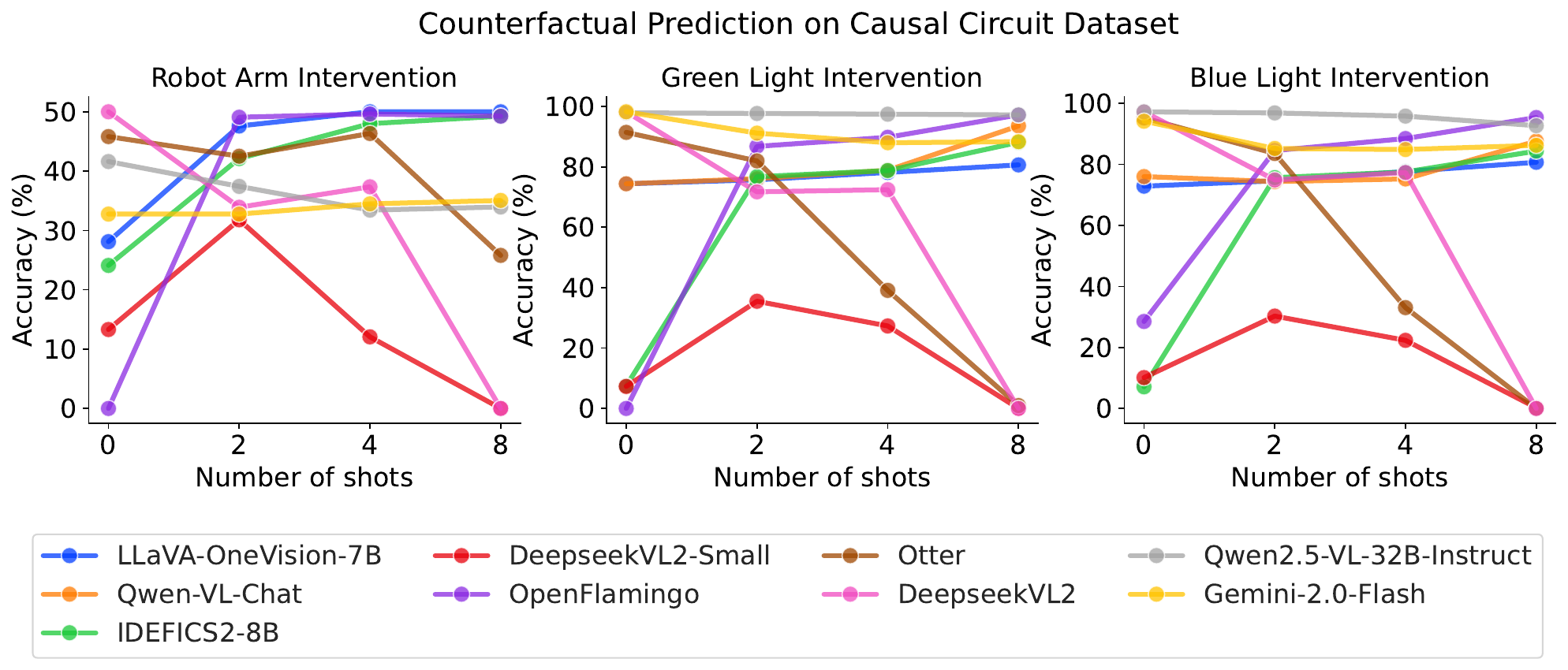}
        \caption{Average Counterfactual Prediction accuracy for causal circuit interventions with at least one descendant.}
        \label{fig:cf_circuit_plot}
    \end{subfigure}

    \caption{Accuracy vs. Shot Count in Counterfactual Inference Tasks across Pendulum, Water Flow, and Causal Circuit datasets for intervened variables with at least one descendant.}
    \label{fig:cf_combined_plot}
\end{figure*}

\paragraph{Per-variable results for Counterfactual Prediction.} Furthermore, in~\cref{tab:counterfactual_pendulum}, ~\cref{tab:counterfactual_flow}, and ~\cref{tab:counterfactual_circuit}, we report the accuracy for samples with the same intervention target to understand how accurately the model predicts all counterfactual states for each variable intervened on separately. We also show results for variables that have at least one descendant and the trend for all models in~\cref{fig:cf_combined_plot}. We observe that most models seem to perform poorly in predicting the accurate counterfactual state, often obtaining around 50\%-60\% in accuracy.

\section{Prompt Templates}
We provide the prompt templates for standard causal structure inference (\cref{fig:discovery_prompts}), interleaved causal structure inference (\cref{fig:interleaved_discovery_prompts}), intervention target prediction (\cref{fig:intervention_prompts}), counterfactual prediction (\cref{fig:counterfactual_prompts}), and chain-of-thought reasoning (\cref{fig:cot_template}, \cref{fig:cot_process}, and \cref{fig:cot_traces}).

\begin{table*}[ht]
  \caption{Configurations of models used in CausalVLBench.}
  \label{tab:models}
  \centering
  \setlength\tabcolsep{6pt}
  
  \resizebox{1.0\textwidth}{!}{ 
  \begin{tabular}{lllcc}
    \toprule
    \textbf{Model} & \textbf{Connection Module} & \textbf{Image Tokens} & \textbf{Context Length (Train)}& \textbf{Context Length (Test)}  \\
    \midrule
    OpenFlamingo-9B  & Perceiver & 64 & 2048 & 2048 \\
    Otter-LLaMA-9B & Perceiver & 64 &  2048 & 2048 \\
    Qwen-VL-Chat-9B & Cross-Attention & 256 & 2048 & 8192  \\
    IDEFICS2-8B & MLP & 64 & - & 32K \\
    LLaVA-OneVision-7B & MLP &  AnyRes & - & 128K\\
    DeepSeekVL2-16B/27B & MLP & AnyRes & 4096 & 128K \\
    Qwen2.5-VL-Instruct-32B & Cross-Attention & 256 & 2048 & 8192  \\
    Gemini-2.0-Flash & Cross-Attention & N/A & N/A & 1M \\
  \bottomrule
  \end{tabular}
  }
\end{table*}

\begin{table*}[t]
    \renewcommand\arraystretch{1.0}
    \centering
    \caption{\textbf{Precision and Recall} Results for \textbf{Task 1A: Standard Causal Structure Inference} and \textbf{Task 1B: Interleaved Causal Structure Inference} task under Zero-Shot setting for selected models}
    \label{tab:prec_rec_csi}
    \setlength{\tabcolsep}{2pt}
    \resizebox{\textwidth}{!}{ 
    \begin{tabular}{lcccccccccccc}
    \toprule
         \multirow{4}{*}{\bf{Model}} & \multicolumn{4}{c}{\textbf{Pendulum}} & \multicolumn{4}{c}{\textbf{Water Flow}} & \multicolumn{4}{c}{\textbf{Causal Circuit}} \\ 
         \cmidrule(lr){2-5} \cmidrule(lr){6-9} \cmidrule(lr){10-13}
         & \multicolumn{2}{c}{\textbf{Standard}} & \multicolumn{2}{c}{\textbf{Interleaved}} 
         & \multicolumn{2}{c}{\textbf{Standard}} & \multicolumn{2}{c}{\textbf{Interleaved}} 
         & \multicolumn{2}{c}{\textbf{Standard}} & \multicolumn{2}{c}{\textbf{Interleaved}} \\
         \cmidrule(lr){2-3}\cmidrule(lr){4-5}\cmidrule(lr){6-7}\cmidrule(lr){8-9}\cmidrule(lr){10-11}\cmidrule(lr){12-13}
         & \textbf{Precision} & \textbf{Recall} & \textbf{Precision} & \textbf{Recall} 
         & \textbf{Precision} & \textbf{Recall} & \textbf{Precision} & \textbf{Recall} 
         & \textbf{Precision} & \textbf{Recall} & \textbf{Precision} & \textbf{Recall} \\ 
         \midrule
         LLaVA-OneVision-7B   & $77.2_{0.0}$ & $100.0_{0.0}$ &  $69.8_{0.2}$ & $100.0_{0.0}$ & $51.6_{0.1}$ & $100.0_{0.0}$  & $50.0_{0.0}$ & $66.7_{0.0}$   & $58.3_{0.8}$ & $29.9_{0.6}$ & $85.2_{0.4}$ & $44.4_{0.1}$   \\
         IDEFICS2-8B      & $100.0_{0.00}$ & $78.9_{0.08}$ & $95.5_{0.11}$ & $100.0_{0.00}$ & $74.7_{0.03}$ & $100.0_{0.03}$ & $50.0_{0.00}$ & $66.7_{0.00}$ & $3.9_{0.05}$ & $1.5_{0.02}$ & $4.2_{0.24}$ & $0.9_{0.04}$   \\
         Qwen2.5-VL-Instruct-32B & $100.0_{0.00}$ & $100.0_{0.00}$  & $100.0_{0.00}$ & $99.9_{0.02}$ & $50.2_{0.05}$ & $100.0_{0.00}$ & $58.4_{0.39}$ & $59.0_{0.66}$ & $99.2_{0.28}$ & $41.3_{0.68}$ & $42.0_{0.99}$ & $8.9_{0.24}$ \\ 
         \midrule
         Gemini-2.0-Flash  & $100.00_{0.00}$ & $100.00_{0.00}$ & $96.8_{0.00}$ & $82.4_{0.00}$ & $74.8_{0.00}$ & $100.0_{0.00}$ & $69.9_{0.00}$ & $52.0_{0.00}$ & $100.0_{0.00}$ & $35.6_{0.00}$ & $99.0_{0.00}$ & $47.8_{0.00}$   \\
    \bottomrule
    \end{tabular}}
\end{table*}

\begin{table*}[htb]
    \renewcommand\arraystretch{1.0}
    \centering
    \caption{\textbf{Bidirectionality vs. Cyclicity} scores for \textbf{Task 1A: Standard Causal Structure Inference}}
    \label{tab:bvc}
    \setlength{\tabcolsep}{4pt}
    \resizebox{\textwidth}{!}{ 
    \begin{tabular}{lcccccc}
    \toprule
         \multirow{2}{*}{\bf{Model}} & \multicolumn{2}{c}{\textbf{Pendulum}} & \multicolumn{2}{c}{\textbf{Water Flow}} & \multicolumn{2}{c}{\textbf{Causal Circuit}} \\ 
         \cmidrule(lr){2-3} \cmidrule(lr){4-5} \cmidrule(lr){6-7}
         & \textbf{Bidirectionality} & \textbf{Cyclicity}
         & \textbf{Bidirectionality} & \textbf{Cyclicity}
         & \textbf{Bidirectionality} & \textbf{Cyclicity} \\ \midrule
         LLaVA-OneVision-7B   & $0.002_{0.000}$ & $0.013_{0.001}$  & $0.167_{0.000}$ & $1.086_{0.000}$  & $0.159_{0.001}$ & $1.053_{0.004}$  \\
         Qwen-VL-Chat-9B       & $0.167_{0.000}$ & $1.086_{0.000}$  & $0.000_{0.000}$ & $0.000_{0.000}$  & $0.000_{0.000}$ & $0.002_{0.001}$  \\ 
         IDEFICS2-8B      & $0.000_{0.000}$ & $0.000_{0.000}$  & $0.167_{0.000}$ & $1.086_{0.000}$  & $0.011_{0.000}$ & $0.074_{0.001}$  \\ 
         Deepseek-VL2-Small-16B       & $0.000_{0.000}$ & $0.000_{0.000}$ & $0.000_{0.000}$ & $0.000_{0.000}$  & $0.000_{0.000}$ & $0.000_{0.000}$  \\ 
         OpenFlamingo-9B       & $0.000_{0.000}$ & $0.000_{0.000}$  & $0.000_{0.000}$ & $0.000_{0.000}$  & $0.000_{0.000}$ & $0.000_{0.000}$  \\ 
         Otter-9B       & $0.000_{0.000}$ & $0.000_{0.000}$  & $0.334_{0.000}$ & $2.966_{0.006}$  & $0.057_{0.003}$ & $0.453_{0.028}$  \\ 
          Deepseek-VL2-27B      & $0.000_{0.000}$ & $0.000_{0.000}$  & $0.000_{0.000}$ & $0.000_{0.000}$  & $0.000_{0.000}$ & $0.000_{0.000}$  \\ 
         Qwen2.5-VL-Instruct-32B       & $0.000_{0.000}$ & $0.000_{0.000}$  & $0.167_{0.000}$ & $1.086_{0.000}$  & $0.000_{0.000}$ & $0.000_{0.000}$  \\ 
         Gemini-Flash-2.0     & $0.000_{0.000}$ & $0.000_{0.000}$  & $0.002_{0.000}$ & $0.011_{0.000}$   & $0.000_{0.000}$ & $0.000_{0.000}$  \\
    \bottomrule
    \end{tabular}}
\end{table*}

\clearpage

\begin{table*}
    \renewcommand\arraystretch{1.0}
    \centering
    \caption{Per-variable results for \textbf{Task 3: Counterfactual Prediction} task under Zero Shot (ZS) and Few Shot (FS) settings for \textbf{Pendulum} dataset \textbf{with causal graph}.}
    \setlength{\tabcolsep}{2pt}
    \resizebox{\textwidth}{!}{ 
    \begin{tabular}{lcccccccccccccccc}
    \toprule
         \multirow{3}{*}{\bf{Model}} & \multicolumn{4}{c}{\textbf{Pendulum Angle}} & \multicolumn{4}{c}{\textbf{Light Position}} & \multicolumn{4}{c}{\textbf{Shadow Length}} & \multicolumn{4}{c}{\textbf{Shadow Position}} \\ 
         \cmidrule(lr){2-5} \cmidrule(lr){6-9} \cmidrule(lr){10-13} \cmidrule(lr){14-17}
         & \textbf{ZS} & \multicolumn{3}{c}{\textbf{FS}} 
         & \textbf{ZS} & \multicolumn{3}{c}{\textbf{FS}} 
         & \textbf{ZS} & \multicolumn{3}{c}{\textbf{FS}} 
         & \textbf{ZS} & \multicolumn{3}{c}{\textbf{FS}} \\ 
         \cmidrule(lr){3-5} \cmidrule(lr){7-9} \cmidrule(lr){11-13} \cmidrule(lr){15-17}
         \textbf{Shots} & \textbf{0} & \textbf{2} & \textbf{4} & \textbf{8} 
         & \textbf{0} & \textbf{2} & \textbf{4} & \textbf{8} 
         & \textbf{0} & \textbf{2} & \textbf{4} & \textbf{8}
         & \textbf{0} & \textbf{2} & \textbf{4} & \textbf{8} \\ \midrule
         LLaVA-OneVision-7B  & $\underline{71.27}$ & $70.25$ & $\underline{70.77}$ & $70.87$ & $65.33$ & $66.73$ & $67.99$ & $69.05$ & $\mathbf{100.00}$ & $99.49$ & $99.52$ & $\underline{99.87}$ & $\mathbf{100.00}$ & $95.49$ & $97.03$ & $\underline{94.92}$ \\
         Qwen-VL-Chat-9B & $65.96$ & $68.12$ & $68.08$ & $62.29$ & $\underline{68.17}$ & $68.51$ & $66.16$ & $55.77$ & $88.18$ & $96.90$ & $96.90$ & $86.84$ & $\underline{98.98}$ & $93.75$ & $90.57$ & $90.03$ \\
         IDEFICS2-8B & $44.96$ & $68.06$ & $69.37$ & $68.88$ & $33.96$ & $64.18$ & $66.62$ & $66.36$ & $35.49$ & $97.93$ & $96.93$ & $98.13$ & $41.02$ & $88.16$ & $88.51$ & $87.84$ \\
         Deepseek-VL2-Small-16B  & $67.84$ & $13.08$ & $0.00$ & $0.00$ & $59.27$ & $21.38$ & $0.00$ & $0.00$ & $\underline{99.49}$ & $27.25$ & $0.00$ & $0.00$ & $84.89$ & $19.45$ & $0.00$ & $0.00$ \\ 
         OpenFlamingo-9B & $0.00$ & $\underline{70.98}$ & $70.25$ & $68.96$ & $7.36$ & $64.24$ & $65.76$ & $67.09$ & $70.58$ & $\underline{99.87}$ & $97.83$ & $99.31$ & $9.64$ & $96.00$ & $93.43$ & $91.36$ \\ 
         Otter-9B & $36.49$ & $7.92$ & $0.07$ & $0.07$ & $25.79$ & $15.38$ & $1.41$ & $0.03$ & $7.23$ & $6.14$ & $0.00$ & $0.07$ & $20.25$ & $24.14$ & $5.27$ & $1.30$ \\ 
         Deepseek-VL2-27B & $\mathbf{71.54}$ & $\mathbf{71.84}$ & $\mathbf{72.04}$ & $0.99$  & $64.72$ & $64.91$ & $65.00$ & $1.70$ & $86.00$ & $\mathbf{100.00}$ & $\mathbf{100.00}$ & $2.18$ & $21.37$ & $\underline{99.90}$ & $\mathbf{100.00}$ & $3.42$ \\ 
         Qwen2.5-VL-Instruct-32B & $64.74$ & $70.31$ & $69.88$ & $\mathbf{73.82}$ & $65.24$ & $\underline{71.64}$ & $\mathbf{73.14}$ & $\mathbf{76.94}$ & $\mathbf{100.00}$ & $\mathbf{100.00}$ & $\underline{99.97}$ & $\mathbf{100.00}$ & $98.63$ & $99.73$ & $\mathbf{100.00}$ & $\mathbf{100.00}$ \\
         \midrule
         Gemini-2.0-Flash   & $69.66$ & $68.87$ & $69.76$ & $\underline{71.44}$ & $\mathbf{72.92}$ & $\mathbf{72.55}$ & $\underline{72.17}$ & $\underline{76.13}$ & $\mathbf{100.00}$ & $\mathbf{100.00}$ & $\mathbf{100.00}$ & $\mathbf{100.00}$ & $92.53$ & $\mathbf{100.00}$ & $\underline{99.90}$ & $\mathbf{100.00}$ \\
    \bottomrule
    \end{tabular}}
    \label{tab:counterfactual_pendulum}
\end{table*}

\begin{table*}[ht]
    \renewcommand\arraystretch{1.0}
    \centering
    \caption{Per-variable results for \textbf{Task 3: Counterfactual Prediction} task under Zero Shot (ZS) and Few Shot (FS) settings for \textbf{Water Flow} dataset \textbf{with causal graph}.}
    \setlength{\tabcolsep}{4pt}
    \resizebox{\textwidth}{!}{ 
    \begin{tabular}{lccccccccccccc}
    \toprule
         \multirow{3}{*}{\bf{Model}} & \multicolumn{4}{c}{\textbf{Ball Size}} & \multicolumn{4}{c}{\textbf{Hole Position}} & \multicolumn{4}{c}{\textbf{Water Level}} \\ 
         \cmidrule(lr){2-5} \cmidrule(lr){6-9} \cmidrule(lr){10-13}
         & \textbf{ZS} & \multicolumn{3}{c}{\textbf{FS}} 
         & \textbf{ZS} & \multicolumn{3}{c}{\textbf{FS}} 
         & \textbf{ZS} & \multicolumn{3}{c}{\textbf{FS}} \\ 
         \cmidrule(lr){3-5} \cmidrule(lr){7-9} \cmidrule(lr){11-13}
         \textbf{Shots} & \textbf{0} & \textbf{2} & \textbf{4} & \textbf{8} 
         & \textbf{0} & \textbf{2} & \textbf{4} & \textbf{8} 
         & \textbf{0} & \textbf{2} & \textbf{4} & \textbf{8} \\ \midrule
         LLaVA-OneVision-7B  & $79.52$ & $\mathbf{81.03}$ & $\mathbf{80.17}$ & $80.44$ & $\underline{85.92}$ & $85.32$ & $85.60$ & $87.54$  & $85.01$ & $85.77$ & $86.89$ & $87.30$  \\
         Qwen-VL-Chat-9B & $\underline{80.19}$ & $\underline{80.34}$ & $\underline{79.89}$ & $78.74$ & $82.55$ & $81.77$ & $82.50$ & $81.06$ & $87.44$ & $84.93$ & $85.51$ & $84.04$ \\
         IDEFICS2-8B & $65.38$ & $79.40$ & $79.25$ & $79.46$ & $72.68$ & $84.26$ & $84.92$ & $85.65$ & $75.00$ & $85.08$ & $87.33$ & $87.51$ \\
         Deepseek-VL2-Small-16B  & $44.88$ & $71.03$ & $7.78$ & $0.00$ & $34.71$ & $60.58$ & $0.20$ & $0.03$ & $74.90$ & $82.20$ & $0.23$ & $0.00$ \\ 
         OpenFlamingo-9B & $20.44$ & $77.60$ & $76.75$ & $\underline{80.66}$ & $46.84$ & $82.09$ & $82.82$ & $85.34$ & $31.74$ & $82.48$ & $83.66$ & $86.51$ \\ 
         Otter-9B & $39.05$ & $41.50$ & $38.33$ & $4.88$ & $17.64$ & $17.23$ & $12.58$ & $2.15$ & $45.36$ & $37.07$ & $24.88$ & $11.30$ \\ 
         Deepseek-VL2-27B & $\mathbf{81.63}$ & $79.85$ & $79.85$ & $0.00$ & $84.16$ & $85.56$ & $86.49$ & $0.00$ & $83.49$ & $84.05$ & $83.57$ & $0.00$ \\ 
         Qwen2.5-VL-Instruct-32B & $61.53$ & $69.09$ & $72.83$ & $77.83$ & $\mathbf{87.22}$ & $\underline{88.49}$ & $\underline{88.83}$ & $\underline{90.01}$ & $\mathbf{91.63}$ & $\mathbf{90.73}$ & $\underline{91.99}$ & $\underline{92.57}$ \\ 
         \midrule
         Gemini-2.0-Flash  & $67.62$ & $75.00$ & $78.21$ & $\mathbf{81.90}$ & $85.71$ & $\mathbf{88.74}$ & $\mathbf{90.30}$ & $\mathbf{90.92}$ & $\underline{89.58}$ & $\underline{90.54}$ & $\mathbf{92.15}$ & $\mathbf{93.19}$   \\
    \bottomrule
    \end{tabular}}
    \label{tab:counterfactual_flow}
\end{table*}

\begin{table*}[ht]
    \renewcommand\arraystretch{1.0}
    \centering
    \caption{Per-variable results for \textbf{Task 3: Counterfactual Prediction} task under Zero Shot (ZS) and Few Shot (FS) settings for \textbf{Causal Circuit} dataset \textbf{with causal graph}.}
    \setlength{\tabcolsep}{2pt}
    \resizebox{\textwidth}{!}{ 
    \begin{tabular}{lcccccccccccccccc}
    \toprule
         \multirow{3}{*}{\bf{Model}} & \multicolumn{4}{c}{\textbf{Robot Arm}} & \multicolumn{4}{c}{\textbf{Green Light}} & \multicolumn{4}{c}{\textbf{Blue Light}} & \multicolumn{4}{c}{\textbf{Red Light}} \\ 
         \cmidrule(lr){2-5} \cmidrule(lr){6-9} \cmidrule(lr){10-13} \cmidrule(lr){14-17}
         & \textbf{ZS} & \multicolumn{3}{c}{\textbf{FS}} 
         & \textbf{ZS} & \multicolumn{3}{c}{\textbf{FS}} 
         & \textbf{ZS} & \multicolumn{3}{c}{\textbf{FS}} 
         & \textbf{ZS} & \multicolumn{3}{c}{\textbf{FS}} \\ 
         \cmidrule(lr){3-5} \cmidrule(lr){7-9} \cmidrule(lr){11-13} \cmidrule(lr){15-17}
         \textbf{Shots} & \textbf{0} & \textbf{2} & \textbf{4} & \textbf{8} 
         & \textbf{0} & \textbf{2} & \textbf{4} & \textbf{8} 
         & \textbf{0} & \textbf{2} & \textbf{4} & \textbf{8}
         & \textbf{0} & \textbf{2} & \textbf{4} & \textbf{8} \\ \midrule
         LLaVA-OneVision-7B & $87.22$ & $\underline{97.25}$ & $\mathbf{99.73}$ & $\mathbf{99.46}$ & $93.60$ & $93.81$ & $94.51$ & $95.10$ & $92.84$ & $93.48$ & $94.36$ & $95.00$ & $99.55$ & $99.14$ & $99.43$ & $99.34$ \\
         Qwen-VL-Chat-9B & $88.60$ & $\mathbf{98.25}$ & $98.17$ & $91.46$ & $68.83$ & $93.00$ & $92.61$ & $95.03$ & $68.54$ & $91.76$ & $90.72$ & $91.16$ & $80.05$ & $97.76$ & $97.91$ & $98.75$ \\
         IDEFICS2-8B & $59.33$ & $94.24$ & $\underline{98.64}$ & $\underline{98.45}$ & $51.87$ & $87.76$ & $92.46$ & $94.66$ & $45.75$ & $89.24$ & $92.75$ & $95.10$ & $73.42$ & $96.80$ & $98.74$ & $99.34$ \\
         Deepseek-VL2-Small-16B  & $28.18$ & $66.41$ & $28.15$ & $0.00$ & $58.91$ & $70.68$ & $24.69$ & $0.00$ & $58.14$ & $70.58$ & $24.09$ & $0.06$ & $17.60$ & $54.82$ & $16.17$ & $0.05$ \\ 
         OpenFlamingo-9B & $0.00$ & $89.85$ & $84.51$ & $89.60$ & $0.00$ & $90.72$ & $95.13$ & $87.84$ & $14.26$ & $84.38$ & $91.96$ & $89.58$ & $0.00$ & $88.03$ & $94.90$ & $89.62$ \\ 
         Otter-9B & $40.06$ & $55.36$ & $55.66$ & $30.49$ & $57.09$ & $60.33$ & $26.91$ & $0.65$ & $53.80$ & $51.01$ & $19.40$ & $0.00$ & $26.55$ & $48.84$ & $56.71$ & $9.12$ \\ 
         Deepseek-VL2-27B & $\mathbf{94.54}$ & $81.32$ & $81.90$ & $1.44$ & $88.96$ & $92.84$ & $93.11$ & $3.80$ & $92.61$ & $93.73$ & $94.24$ & $1.98$ & $99.78$ & $98.38$ & $\underline{99.93}$ & $0.74$ \\ 
         Qwen2.5-VL-Instruct-32B & $\underline{93.53}$ & $91.43$ & $89.89$ & $90.23$ & $\mathbf{99.48}$ & $\mathbf{99.42}$ & $\mathbf{99.36}$ & $\mathbf{99.30}$ & $\mathbf{99.31}$ & $\mathbf{99.22}$ & $\mathbf{98.96}$ & $\mathbf{98.17}$ & $\mathbf{100.00}$ & $\mathbf{100.00}$ & $\mathbf{100.00}$ & $\mathbf{100.00}$ \\
         \midrule
         Gemini-2.0-Flash & $91.09$ & $91.38$ & $92.24$ & $92.53$ & $\underline{98.41}$ & $\underline{97.79}$ & $\underline{97.00}$ & $\underline{97.08}$ & $\underline{93.99}$ & $\underline{96.22}$ & $\underline{96.05}$ & $\underline{96.56}$ & $\underline{99.78}$ & $\underline{99.71}$ & $99.48$ & $\underline{99.56}$ \\
    \bottomrule
    \end{tabular}}
    \label{tab:counterfactual_circuit}
\end{table*}

\newpage

\clearpage
\begin{figure*}
    \centering
    \begin{subfigure}[t]{\textwidth}
        \centering
    \begin{tcolorbox}[colback=gray!10,left=2pt, right=2pt, top=2pt, bottom=2pt, fontupper=\small, fontlower=\small]
\textbf{Standard Causal Structure Inference - Pendulum}

\textbf{Task Instruction:}
You are a highly capable AI system specialized in causal reasoning from visual data. You will be shown an image containing a physical setup with a light source, a pendulum, and the pendulum’s shadow. The scene contains four variables that are causally related: pendulum angle, light position, shadow length, and shadow position. Given an image and a question about two variables, A and B, your task is to determine whether A causes B. Answer simply with Yes or No.
 \\\\
    \textbf{Query:} 
<image> Does [variable A] directly cause [variable B] to change?
    \end{tcolorbox}
    \end{subfigure} \\
    \begin{subfigure}[t]{\textwidth}
        \centering
    \begin{tcolorbox}[colback=gray!10,left=2pt, right=2pt, top=2pt, bottom=2pt, fontupper=\small, fontlower=\small]
\textbf{Standard Causal Structure Inference - Water Flow}

\textbf{Task Instruction:}
You are a highly capable AI system specialized in causal reasoning from visual data. You will be shown an image containing a physical setup with water in a glass and a hole on the right side of the glass from where the water is flowing. There is also a red ball inside the glass that affects the water level in the glass and the water flow from the hole. The scene contains four variables that are causally related: ball size, water level, hole position, and water flow. Given an image and a question about two variables, A and B, your task is to determine whether A causes B. Answer simply with Yes or No.
 \\\\
    \textbf{Query:} 
<image> Does [variable A] directly cause [variable B] to change?
    \end{tcolorbox}
    
    \end{subfigure}\\
    \begin{subfigure}[t]{\textwidth}
        \centering
    \begin{tcolorbox}[colback=gray!10,left=2pt, right=2pt, top=2pt, bottom=2pt, fontupper=\small, fontlower=\small]
\textbf{Standard Causal Structure Inference - Causal Circuit}

\textbf{Task Instruction:}
You are a highly capable AI system specialized in causal reasoning from visual data. You will be shown an image containing a physical setup showing a robotic arm positioned over a circular arc with three buttons (red, green, blue) and the resulting lighting in the scene. The scene contains four variables that are causally related: robot arm, green light, blue light, and red light. Given an image and a question about two variables, A and B, your task is to determine whether A causes B. Answer simply with Yes or No.
 \\\\
    \textbf{Query:} 
<image> Does [variable A] directly cause [variable B] to change?
    \end{tcolorbox}
    \end{subfigure}
    \caption{Prompt templates for \textbf{Task 1A: Standard Causal Structure Inference} task}
    \label{fig:discovery_prompts}
\end{figure*}

\begin{figure*}
    \centering
    \begin{subfigure}[t]{\textwidth}
        \centering
    \begin{tcolorbox}[colback=gray!10,left=2pt, right=2pt, top=2pt, bottom=2pt, fontupper=\small, fontlower=\small]
\textbf{Interleaved Causal Structure Inference - Pendulum}

\textbf{Task Instruction:}
You are a highly capable AI system specialized in causal reasoning from visual data. You will be shown two images: the first image shows a physical setup with a light source, a pendulum, and the pendulum’s shadow. The scene contains four variables that are causally related: pendulum angle, light position, shadow length, and shadow position. The second image shows the same setup after one of these variables is initially changed and other variables may have changed as a downstream effect. Given a pair of images and a question about two variables, A and B, your task is to determine whether A causes B. Answer simply with Yes or No.
 \\\\
    \textbf{Query:} 
<image> Does [variable A] directly cause [variable B] to change?
    \end{tcolorbox}
    \end{subfigure} \\
    \begin{subfigure}[t]{\textwidth}
        \centering
    \begin{tcolorbox}[colback=gray!10,left=2pt, right=2pt, top=2pt, bottom=2pt, fontupper=\small, fontlower=\small]
\textbf{Interleaved Causal Structure Inference - Water Flow}

\textbf{Task Instruction:}
You are a highly capable AI system specialized in causal reasoning from visual data. You will be shown two images: the first image shows a physical setup with water in a glass, a hole on the right side of the glass from where the water is flowing, and a red ball inside the glass. The scene contains four variables that are causally related: ball size, hole position, water level, and water flow. The second image shows the same setup after one of these variables is initially changed and other variables may have changed as a downstream effect. Given a pair of images and a question about two variables, A and B, your task is to determine whether A causes B. Answer simply with Yes or No.
 \\\\
    \textbf{Query:} 
<image> Does [variable A] directly cause [variable B] to change?
    \end{tcolorbox}
    \end{subfigure}\\
    \begin{subfigure}[t]{\textwidth}
        \centering
    \begin{tcolorbox}[colback=gray!10,left=2pt, right=2pt, top=2pt, bottom=2pt, fontupper=\small, fontlower=\small]
\textbf{Interleaved Causal Structure Inference - Causal Circuit}

\textbf{Task Instruction:}
You are a highly capable AI system specialized in causal reasoning from visual data. You are given two images of a robotic scene. The first image shows a robotic arm positioned over a circular arc with three buttons (red, green, blue) and the resulting lighting in the scene. The scene contains four variables that are causally related: robot arm, green light, blue light, and red light. The second image shows the same setup after one of these variables is initially changed and other variables may have changed as a downstream effect. Given a pair of images and a question about two variables, A and B, your task is to determine whether A causes B. Answer simply with Yes or No.
 \\\\
    \textbf{Query:} 
<image> Does [variable A] directly cause [variable B] to change?
    \end{tcolorbox}
    \end{subfigure}
    \caption{Prompt templates for \textbf{Task 1B: Interleaved Causal Structure Inference} task}
    \label{fig:interleaved_discovery_prompts}
\end{figure*}

\begin{figure*}
    \centering
    \begin{subfigure}[t]{\textwidth}
        \centering
    \begin{tcolorbox}[colback=gray!10,left=2pt, right=2pt, top=2pt, bottom=2pt, fontupper=\small, fontlower=\small]
\textbf{Intervention Target Prediction - Pendulum}

\textbf{Task Instruction:}
You are a highly capable AI system specialized in causal reasoning from visual data. You will be shown two images: the first image shows a physical setup with a light source, a pendulum, and the pendulum’s shadow. The second image shows the same setup after a change has occurred. The scene contains four variables: pendulum angle, light position, shadow length, and shadow position. These variables are causally related as follows:
\\\\    
(1) If the pendulum angle changes, it causes both the shadow length and shadow position to change. It does NOT cause the light position to change.
    
(2) If the light position changes, it causes both the shadow length and shadow position to change. It does NOT cause the pendulum angle to change.
    
(3) A change in shadow length does NOT cause any other variable to change.
    
(4) A change in shadow position does NOT cause any other variable to change. 
\\\\
Your task is to compare the two images, identify the first variable that changed, and use the causal rules above to determine which variable is the likely root cause of any other changes. Respond with only one of the following variable names, exactly as written: pendulum angle, light position, shadow length, or shadow position.
 \\\\
    \textbf{Query:} 
<image><image> From the first to the second image, which variable changes first?
    \end{tcolorbox}
    
    \end{subfigure} \\
    \begin{subfigure}[t]{\textwidth}
        \centering
    \begin{tcolorbox}[colback=gray!10,left=2pt, right=2pt, top=2pt, bottom=2pt, fontupper=\small, fontlower=\small]
\textbf{Intervention Target Prediction - Water Flow}

\textbf{Task Instruction:}
You are a highly capable AI system specialized in causal reasoning from visual data. You will be shown two images. The first image shows a physical setup with water in a glass and a hole on the right side of the glass from where the water is flowing. There is also a red ball inside the glass that affects the water level in the glass and the water flow from the hole. The second image shows the same setup after a change has occurred. The scene contains four variables: ball size, water level, hole position, and water flow. These variables are causally related as follows:
\\\\
(1) If the ball size changes, it causes the water level to change and affects water flow. It does NOT cause hole position to change.
    
(2) If the water level changes, it causes the water flow to change. It does NOT cause ball size and hole position to change.
    
(3) If the hole position changes, it causes water flow to change. It does NOT cause ball size or water level to change.
\\\\
Your task is to compare the two images, identify the first variable that changed, and use the causal rules above to determine which variable is the likely root cause of any other changes. Respond with only one of the following variable names, exactly as written: ball size, water level, hole position.
 \\\\
    \textbf{Query:} 
<image><image> From the first to the second image, which variable changes first?
    \end{tcolorbox}
    
    \end{subfigure}\\
    \begin{subfigure}[t]{\textwidth}
        \centering
    \begin{tcolorbox}[colback=gray!10,left=2pt, right=2pt, top=2pt, bottom=2pt, fontupper=\small, fontlower=\small]
\textbf{Intervention Target Prediction - Causal Circuit}

\textbf{Task Instruction:}
You are a highly capable AI system specialized in causal reasoning from visual data. You are given two images of a robotic scene: a before image and an after image. Each image shows a robotic arm positioned over a circular arc with three buttons (red, green, blue) and the resulting lighting in the scene. The scene contains four variables: robot arm, green light, blue light, and red light. These variables are causally related as follows:
\\\\
(1) Changing the arm position causes one button to be pressed, which directly affects the corresponding light (red, green, or blue).
    
(2) Turning on the green or blue light can indirectly activate the Red light.
    
(3) Changing any light does not affect the arm position.
\\\\
Your task is to compare the two images, identify the first variable that changed, and use the causal rules above to determine which variable is the likely root cause of any other changes. Respond with only one of the following variable names, exactly as written: robot arm, green light, blue light, red light.
 \\\\
    \textbf{Query:} 
<image><image> From the first to the second image, which variable changes first?
    \end{tcolorbox}
    
    \end{subfigure}
    \caption{Prompt templates for \textbf{Task 2: Intervention Target Prediction} task}
    \label{fig:intervention_prompts}
\end{figure*}

\begin{figure*}
    \centering
    \begin{subfigure}[t]{\textwidth}
        \centering
    \begin{tcolorbox}[colback=gray!10,left=2pt, right=2pt, top=2pt, bottom=2pt, fontupper=\small, fontlower=\small]
\textbf{Counterfactual Prediction - Pendulum}

\textbf{Task Instruction:}
You are a highly capable AI system specialized in causal reasoning from visual data. You will be shown an image containing a physical setup with a light source, a pendulum, and the pendulum’s shadow. The scene contains four variables: pendulum angle, light position, shadow length, and shadow position. The pendulum angle can be one of the following values: left, center, right. The light position can be one of the following values: right, center, left. The shadow length can be one of the following values: short, medium, long. The shadow position can be one of the following values: left, center, right. These variables are causally related as follows:
 \\\\
(1) If the pendulum angle changes, it causes both the shadow length and shadow position to change. It does NOT cause the light position to change.
    
(2) If the light position changes, it causes both the shadow length and shadow position to change. It does NOT cause the pendulum angle to change.
    
(3) A change in shadow length does NOT cause any other variable to change.
    
(4) A change in shadow position does NOT cause any other variable to change.
 \\\\
Given an image and a variable that will change, your task is to determine what the final values of all four variables would be had the variable been changed to the specified value.
 \\
    \textbf{Query:} 
<image> <image> In the given image, the values of the variables are given as pendulum angle: \{\}, light position: \{\}, shadow length: \{\}, shadow position: \{\}
 \\\\
If the [intervened variable] had been changed from x to y, what would be the final values of all variables? Answer concisely with the specific values that each variable will take.
    \end{tcolorbox}
    
    \end{subfigure} \\
    \begin{subfigure}[t]{\textwidth}
        \centering
    \begin{tcolorbox}[colback=gray!10,left=2pt, right=2pt, top=2pt, bottom=2pt, fontupper=\small, fontlower=\small]
\textbf{Counterfactual Prediction - Water Flow}

\textbf{Task Instruction:}
You are a highly capable AI system specialized in causal reasoning from visual data. You will be shown an image containing a physical setup with water in a glass and a hole on the right side of the glass from where the water is flowing. There is also a red ball inside the glass that affects the water level in the glass and the water flow from the hole. The scene contains four variables: ball size, water level, hole position, and water flow. The ball size can be one of the following values: small, medium, large. The hole position can be one of the following values: bottom, middle, top. The water level can be one of the following values: low, medium, high. The water flow can be one of the following values: left, middle, right. For water level, left refers to close to the glass and right refers to far from the glass. These variables are causally related as follows:
 \\\\    
(1) If the ball size changes, it causes the water level to change and affects water flow. It does NOT cause hole position to change.
    
(2) If the water level changes, it causes the water flow to change. It does NOT cause ball size and hole position to change.
    
(3) If the hole position changes, it causes water flow to change. It does NOT cause ball size or water level to change. 
 \\\\    
Given an image and a variable that will change, your task is to determine what the final values of all four variables would be had the variable been changed to the specified value.
 \\
    \textbf{Query:} 
<image> In the given image, the values of the variables are given as ball size: \{\}, hole position: \{\}, water level: \{\}, water flow: \{\}
 \\\\    
If the [intervened variable] had been changed from x to y, what would be the final values of all variables? Answer concisely with the specific values that each variable will take.
    \end{tcolorbox}
    
    \end{subfigure}\\
    \begin{subfigure}[t]{\textwidth}
        \centering
    \begin{tcolorbox}[colback=gray!10,left=2pt, right=2pt, top=2pt, bottom=2pt, fontupper=\small, fontlower=\small]
\textbf{Counterfactual Prediction - Causal Circuit}

\textbf{Task Instruction:}
You are a highly capable AI system specialized in causal reasoning from visual data. You will be shown an image containing a physical setup showing a robotic arm positioned over a circular arc with three buttons (red, green, blue) and the resulting lighting in the scene. The scene contains four variables: robot arm, green light, blue light, and red light. The robot arm can be one of the following values: touching red light, touching blue light, touching green light, or not touching any light. The red light can be one of the following values: on or off. The green light can be one of the following values: on or off. The blue light can be one of the following values: on or off. These variables are causally related as follows:
\\\\   
(1) Changing the arm position causes one button to be pressed, which directly affects the corresponding light (red, green, or blue).
    
(2) Turning on the green or blue light can indirectly activate the Red light.
    
(3) Changing any light does not affect the arm position.
\\\\   
Given an image and a variable that will change, your task is to determine what the final values of all four variables would be had the variable been changed to the specified value.
 \\
    \textbf{Query:} 
<image> In the given image, the values of the variables are given as red light: \{\}, green light: \{\}, blue light: \{\}, robot arm: \{\}
  \\\\   
If the [intervened variable] had been changed from x to y, what would be the final values of all variables? Answer concisely with the specific values that each variable will take.
    \end{tcolorbox}
    
    \end{subfigure}
    \caption{Prompt templates for \textbf{Task 3: Counterfactual Prediction} task}
    \label{fig:counterfactual_prompts}
\end{figure*}

\begin{figure*}
    \centering
    \begin{tcolorbox}[colback=gray!10,left=2pt, right=2pt, top=2pt, bottom=2pt, fontupper=\small, fontlower=\small]
    \textbf{Standard Causal Structure Inference CoT Prompt} \\
    Let's think step by step. First, analyze the location of all objects in the image. Then, determine the relationships between the variables. Give reasoning rationales. \\

    \textbf{Interleaved Causal Structure Inference CoT Prompt} \\
    Let's think step by step. First, analyze the location of all objects in the first image. Second, analyze the location of all objects in the second image. Then, determine which variables have been changed according to the rules provided. Finally, determine the relationships between the variables. Give reasoning rationales. \\

    \textbf{Intervention Target Prediction CoT Prompt} \\
    Let's think step by step. First, analyze the location of all objects in the first image. Second, analyze the location of all objects in the second image. Then, determine which variables have been changed according to the rules provided. Give reasoning rationales. \\

    \textbf{Counterfactual Prediction CoT Prompt} \\
    Let's think step by step. First, analyze the location of all objects in the image. Then, determine how each variable would change based on the desired manipulation according to the rules provided. Give reasoning rationales.
    \end{tcolorbox}
    \vspace{-0.48cm}
    \caption{\textbf{Chain-of-Thought} Prompt templates}
    \label{fig:cot_template}
\end{figure*}

\begin{figure*}
    \centering
    \begin{tcolorbox}[colback=gray!10,left=2pt, right=2pt, top=2pt, bottom=2pt, fontupper=\small, fontlower=\small]
\textbf{User:} [Task Description][CoT Prompt]
    
\textbf{LVLM:} [Reasoning Chain]

\textbf{User:} [Task Description][CoT Prompt][Reasoning Chain][Query]

\textbf{LVLM:} [Prediction]
\end{tcolorbox}
    \vspace{-0.48cm}
    \caption{\textbf{Chain-of-Thought} prompting strategy}
    \label{fig:cot_process}
\end{figure*}

\begin{figure*}
    \centering
    \begin{subfigure}[t]{\textwidth}
        \centering
    \begin{tcolorbox}[colback=gray!10,left=2pt, right=2pt, top=2pt, bottom=2pt, fontupper=\small, fontlower=\small]
\textbf{Standard Causal Structure Inference - Pendulum}

1. **Object Locations:** The image shows a light source at the top, a pendulum consisting of a line and a circle in the middle, and the pendulum's shadow at the bottom.

2. **Variable Relationships:**
   * **Pendulum Angle -> Shadow Length:** The angle of the pendulum directly influences the length of its shadow. A larger pendulum angle will generally result in a longer shadow (depending on the light's position).
   
   * **Pendulum Angle -> Shadow Position:** Similarly, the pendulum angle affects the position of the shadow. As the pendulum swings, the shadow's location shifts.
   
   * **Light Position -> Shadow Length:** The position of the light source also plays a role. If the light is directly overhead, the shadow will be short. If the light is at an angle, the shadow will be longer.
   
   * **Light Position -> Shadow Position:** The light position also influences where the shadow falls.

Based on these relationships, the pendulum angle does cause the shadow length.
    \end{tcolorbox}
    
    \end{subfigure} \\
    \begin{subfigure}[t]{\textwidth}
        \centering
    \begin{tcolorbox}[colback=gray!10,left=2pt, right=2pt, top=2pt, bottom=2pt, fontupper=\small, fontlower=\small]
\textbf{Intervention Target Prediction - Pendulum}

In the first image, the pendulum angle is at a certain position, the light position is fixed, and the shadow has a particular length and position. In the second image, the pendulum angle is at the same position, the light position is also the same, the shadow length is the same, but the shadow position is different. Since the pendulum angle and light position are unchanged, the change in shadow position cannot be attributed to them. According to rule (4), a change in shadow position does not cause any other variable to change. Therefore, the initial change must have been in the shadow position.
    \end{tcolorbox}
    
    \end{subfigure}\\
    \begin{subfigure}[t]{\textwidth}
        \centering
    \begin{tcolorbox}[colback=gray!10,left=2pt, right=2pt, top=2pt, bottom=2pt, fontupper=\small, fontlower=\small]
\textbf{Counterfactual Prediction - Pendulum}

Here's a step-by-step analysis:

**1. Initial State:**

*   **Pendulum Angle:** left

*   **Light Position:** left

*   **Shadow Length:** short

*   **Shadow Position:** right

**2. Desired Manipulation:**

We want to change the light position to `right'.

**3. Causal Reasoning:**

*   The rule (2) says that if the light position changes, it causes both the shadow length and shadow position to change. It does NOT cause the pendulum angle to change.

*   Given that the light moves from left to right, the shadow position will move from right to left.

*   Given that the light moves from left to right, the shadow length will become long.

**4. Predicted Final State:**

*   **Pendulum Angle:** left (unchanged)

*   **Light Position:** right

*   **Shadow Length:** long

*   **Shadow Position:** left
    \end{tcolorbox}
    \end{subfigure}
    \vspace{-0.48cm}
    \caption{Gemini-2.0-Flash intermediate \textbf{Chain-of-Thought Reasoning Traces} for Causal Structure Inference, Intervention Target Prediction, and Counterfactual Prediction tasks}
    \label{fig:cot_traces}
\end{figure*}

\end{document}